%% file: main.tex
\documentclass[a4paper]{article}

\pdfoutput=1
\usepackage[utf8]{inputenc}

\usepackage[super,sort&compress,comma]{natbib}

\usepackage[english]{babel}
\usepackage[T1]{fontenc}
\usepackage{csquotes}

\usepackage[a4paper,top=3cm,bottom=2cm,left=3cm,right=3cm,marginparwidth=1.75cm]{geometry}

\usepackage{amsmath}
\usepackage{graphicx}
\usepackage{hyperref}
\hypersetup{colorlinks=true, allcolors=blue}
\usepackage{authblk}
\usepackage{multirow}
\usepackage{float}
\usepackage{natbib}

\usepackage{nomencl}
\usepackage{booktabs}
\usepackage{tabularx}
\usepackage{textcomp} 
\usepackage{amsfonts}
\usepackage{comment} 
\usepackage{outlines} 
\usepackage{subfig}
\usepackage[version=4]{mhchem}
\usepackage{siunitx}
\usepackage{bm}
\usepackage[normalem]{ulem} 
\date{}

\DeclareSIUnit\angstrom{\text {Å}}

\usepackage[final,commandnameprefix=always]{changes}

\title{Multi-stage Bayesian optimisation for dynamic decision-making in self-driving labs}

\author[1,2]{Luca Torresi}
\author[1,2,*]{Pascal Friederich}

\affil[1]{Institute of Nanotechnology, Karlsruhe Institute of Technology, Kaiserstr. 12, 76131 Karlsruhe, Germany}
\affil[2]{Institute of Anthropomatics and Robotics, Karlsruhe Institute of Technology, Kaiserstr. 12, 76131 Karlsruhe, Germany}
\affil[*]{Corresponding author: pascal.friederich@kit.edu}

\begin{document}

\maketitle
\begin{abstract}
    \input{00_abstract.tex}
\end{abstract}

\section{Introduction}\label{sec:introduction}
\input{01_introduction.tex}

\section{Preliminaries}
\input{02_preliminaries.tex}

\section{Method}
\input{03_methods.tex}

\section{Experiments}\label{experiments}
\input{04_experiments.tex}

\section{Conclusion}
\input{05_conclusion.tex}

\subsection*{Code availability}
\input{07_code.tex}

\subsection*{Acknowledgements}
\input{10_acknowledgements.tex}

\printnomenclature

\bibliography{bib}
\bibliographystyle{rsc} 

\clearpage
\setcounter{figure}{0}
\setcounter{table}{0}
\renewcommand{\thefigure}{S\arabic{figure}}
\renewcommand{\thetable}{S\arabic{table}}
\appendix
\section*{Supplementary Information}
\input{appendix.tex}

\end{document}

%% file: 00_abstract.tex
Self-driving laboratories (SDLs) are combining recent technological advances in robotics, automation, and machine learning based data analysis and decision-making to perform autonomous experimentation toward human-directed goals without requiring any direct human intervention.
SDLs are successfully used in materials science, chemistry, and beyond, to optimise processes, materials, and devices in a systematic and data-efficient way.
At present, the most widely used algorithm to identify the most informative next experiment is Bayesian optimisation.
While relatively simple to apply to a wide range of optimisation problems, standard Bayesian optimisation relies on a fixed experimental workflow with a clear set of optimisation parameters and one or more measurable objective functions.
This excludes the possibility of making on-the-fly decisions about changes in the planned sequence of operations and including intermediate measurements in the decision-making process.
Therefore, many real-world experiments need to be adapted and simplified to be converted to the common setting in self-driving labs.
In this paper, we introduce an extension to Bayesian optimisation that allows flexible sampling of multi-stage workflows and makes optimal decisions based on intermediate observables, which we call proxy measurements.
We systematically compare the advantage of taking into account proxy measurements over conventional Bayesian optimisation, in which only the final measurement is observed.
We find that over a wide range of scenarios, proxy measurements yield a substantial improvement, both in the time to find good solutions and in the overall optimality of found solutions.
This not only paves the way to use more complex and thus more realistic experimental workflows in autonomous labs but also to smoothly combine simulations and experiments in the next generation of SDLs.

%% file: 01_introduction.tex
Research in materials science and chemistry has typically relied on intuition and serendipity, and has thus been limited by human cognitive capacity \cite{kitano2021nobel}. The traditional process of identifying efficient materials from a range of imaginable compositions usually spans decades. To move beyond these constraints, the scientific community adopted paradigms such as high-throughput methods \cite{green2013applications} and design of experiments \cite{mongomery2017montgomery}, ultimately evolving the field into the era of self-driving laboratories (SDLs). These systems integrate robotics and machine learning (ML) to enable autonomous experimentation toward human-directed goals without requiring any direct human intervention \cite{abolhasani2023rise, maffettone2023missing}. SDLs perform experiments iteratively, choosing each successive experiment based on previously collected data rather than pre-planning the entire campaign, which maximises the information gain per sample and requires fewer experiments compared to traditional design of experiments \cite{delgado2023research}.

Several ML algorithms have been deployed to drive SDLs, including genetic algorithms \cite{salley2020nanomaterials}, reinforcement learning \cite{rajak2021autonomous, maffettone2021gaming}, and Bayesian optimisation (BO) \cite{macleod2020self}. Among these, BO is the most widely adopted method thanks to its ability to operate effectively with a limited number of evaluations, navigating noisy data and complex landscapes without the need for gradient information. BO has been successfully applied to SDLs for both the optimisation of single objectives \cite{macleod2020self, gongora2020bayesian, shields2021bayesian}, and, more recently, for the simultaneous optimisation of multiple objectives \cite{macleod2022self}. 

Despite the considerable advantages these methods offer over traditional approaches, they also come with inherent limitations. A key shortcoming of all these frameworks is their reliance on a fixed experimental workflow. This rigidity prevents making on-the-fly adjustments to the planned sequence of operations. Additionally, standard approaches typically treat experiments as black boxes, ignoring the known structure of the problem, i.e. the sequential dependencies between process stages and their intermediate results. As a result, existing frameworks often prove too rigid for practical materials science experiments, which are rarely isolated black-box problems but rather complex, multi-stage processes with tunable parameters at each stage and measurable intermediate outcomes.

One illustrative example is the fabrication of perovskite solar cells, which involves multiple separate stages, such as the deposition of the electrode, transport, and perovskite layers \cite{agresti2024scalable}. Although the quality of each layer critically impacts the final device performance, key metrics such as power conversion efficiency can only be measured after the full device is completed. While it is possible to optimise each layer individually \cite{wu2024inverse}, this approach fails to capture the complex interdependencies between the different components.

\paragraph{Related work.}  
Multi-fidelity Bayesian optimisation \cite{lam2015multifidelity, song2019general}, also known as multi-information source BO \cite{poloczek2017multi}, is a well-established strategy that mitigates the high cost and sample inefficiency of fixed workflows. The approach exploits lower-fidelity approximations of the main objective, which are cheaper to evaluate, may be sampled independently, and are correlated with high-fidelity results. These methods model evaluations at different fidelities jointly (e.g., via multi-output Gaussian processes) and strategically balance low-cost approximate measurements with high-cost accurate measurements. Selection strategies either decouple the choice of fidelity and query location in the input parameter space or incorporate evaluation cost directly into the acquisition function to decide both simultaneously. However, MFBO still inherits key limitations of standard BO, including reliance on fixed workflows, limited use of intermediate measurements, and inadequate handling of structured multi-stage processes.

Going beyond multi-fidelity BO, \citet{astudillo2019bayesian} extended BO to structured objectives by considering composite functions of the form $f(x) = g(h(x))$, where $h$ is an expensive vector-valued black box and $g$ is a cheap real-valued function.
They modelled $h$ with a Gaussian process (GP) and designed a Monte Carlo-based acquisition function, called expected improvement for composite functions, which uses intermediate measurements to guide the optimisation process. This framework was later generalised to directed acyclic graphs (DAGs) of functions \cite{astudillo2021bayesian}, where each node depends on its parents and is modelled as an independent GP. This approach, termed Bayesian optimisation of function networks (BOFN), leverages known structure for more efficient optimisation but requires evaluating the full network at each iteration, limiting flexibility in selectively sampling sub-graphs.

Recent extensions, such as partially observable knowledge-gradient methods for function networks~\cite{buathong2023bayesian}, relax the rigidity of standard BO by allowing selective evaluation of intermediate nodes. While the original formulation achieves strong query efficiency, it comes at a high computational cost due to its nested Monte Carlo estimators and repeated acquisition optimisations. A faster variant~\cite{buathong2025fast} reduces runtime by optimising over subsets of the input space, with modest reductions in solution quality. Both approaches assume that intermediate nodes can be probed at arbitrary inputs, independent of upstream stages. While this abstraction facilitates algorithm design, it may not hold in practical laboratory settings, where intermediate measurements usually require executing preceding stages of the process, and the possible value ranges of observations at each stage are typically not known in advance.

Building on cascade Bayesian optimisation~\cite{nguyen2016cascade} and BOFN, \citet{kusakawa2022bayesian} proposed a method for cascade processes, modelling each stage as a separate GP and dynamically adjusting downstream stage parameters conditional on the measured outputs of earlier stages, with the core contribution focusing on this continuous, multi-stage look-ahead sampling. The effectiveness of the method was demonstrated using a combination of synthetic numerical test functions and a solar cell simulator learned from data.

\paragraph{Contributions.}

In this work, we develop a multi-stage Bayesian optimisation framework (MSBO), which integrates proxy and intermediate-stage measurements into BO, enabling flexible, on-the-fly workflow adaptation that reduces optimisation costs.
We design a synthetic function generator for multi-stage optimisation tasks, supporting systematic exploration of varying stage complexities, correlations, and levels of information loss through additive noise and masking. The function generator aims to mimic real-world applications' complexity better than standard analytical test functions used frequently to benchmark BO algorithms.
We extensively evaluate MSBO, showing that integrating a flexible acquisition function with a multi-stage GP surrogate consistently outperforms standard BO across diverse synthetic benchmarks and real-world problems.
We furthermore quantify the efficiency gains in various relevant self-driving lab scenarios involving two-stage processes with varying complexities and cost ratios between the first proxy measurement and the second, more expensive measurement of the objective. We demonstrate that even for the combination of very simple proxy measurements and very complex objective measurements, the multi-stage Bayesian optimisation approach outperforms standard BO, paving the way to using proxy measurements as a default in all self-driving lab settings, and also to including (cheap and approximative) computational steps in experimental optimisation workflows.

%% file: 02_preliminaries.tex
\subsection{Bayesian optimisation}
Bayesian optimisation is an effective framework for finding the global maximum of an expensive black-box function $f(\boldsymbol{x})$ over a bounded domain $\mathcal{X}$:
\begin{equation}
\max_{\boldsymbol{x} \in \mathcal{X}} f(\boldsymbol{x})
\end{equation}
BO is particularly suited for functions where evaluations are costly, noisy, or lack an analytic expression, precluding the use of gradient-based methods \cite{brochu2010tutorial}. The core idea is to balance exploration, i.e. sampling regions with high uncertainty, and exploitation, i.e. sampling regions likely to improve the current best-observed value, to reach the global optimum in a minimal number of function evaluations.

The BO process is iterative, relying on two essential components: a surrogate model and an acquisition function. The surrogate model is a probabilistic model, constructed using a small set of observations to approximate $f(\boldsymbol{x})$. It provides fast estimates of predictions and associated uncertainty across the optimisation domain $\mathcal{X}$. The acquisition function quantifies the expected utility of evaluating $f$ at a specific point $\boldsymbol{x}$. The next sampling location $\boldsymbol{x}_{t+1}$ is chosen by maximising this computationally inexpensive acquisition function.
The alternating execution of these two steps efficiently generates a sequence of samples converging to the global optimum.

\subsubsection{Gaussian processes}
Gaussian processes are a common choice for the BO surrogate model due to their non-parametric nature and reliable, built-in uncertainty quantification \cite{ober2021promises}. A GP is a collection of random variables, any finite number of which have a joint multivariate Gaussian distribution \cite{rasmussen2006gaussian}. It defines a prior distribution over functions $f: \mathcal{X} \rightarrow \mathbb{R}$, specified completely by a mean function $m(\boldsymbol{x}) = \mathbb{E} [f(\boldsymbol{x})]$ and a covariance function $k(\boldsymbol{x}, \boldsymbol{x}') = \mathbb{E}[(f(\boldsymbol{x}) - m(\boldsymbol{x}))(f(\boldsymbol{x}') - m(\boldsymbol{x}'))]$:
\begin{equation}
f(\boldsymbol{x}) \sim \mathcal{G}\mathcal{P}(m(\boldsymbol{x}), k(\boldsymbol{x},\boldsymbol{x}')).
\end{equation}
The covariance function, or kernel $k(\cdot, \cdot)$, is a positive semi-definite function (e.g., Matern or radial basis function kernels) that defines the similarity between input pairs $\boldsymbol{x}$ and $\boldsymbol{x}'$. The mean function is often taken to be zero for simplicity.

While standard GP training suffers from $\mathcal{O}(N^3)$ computational complexity, various approximation methods exist to scale GPs to large datasets \cite{liu2020gaussian}. Furthermore, GPs can be extended to model multiple correlated outputs using multi-task Gaussian processes~\cite{bonilla2007multi}, often employing an intrinsic coregionalization model to define the inter-task covariance structure~\cite{alvarez2012kernels}.

\subsubsection{Acquisition functions}
Acquisition functions $\mathcal{L}(\boldsymbol{x})$ translate the probabilistic belief from the surrogate model into a metric for selecting the next query point. Maximising $\mathcal{L}(\boldsymbol{x})$ replaces the original expensive optimisation problem with a computationally cheap one.

Acquisition functions are typically defined as the expectation of a utility function $l(y)$, which measures the desirability of a hypothetical outcome $y$ given the current input $\boldsymbol{x}$. If $p(y|\boldsymbol{x})$ is the predictive distribution from the surrogate model, $\mathcal{L}(\boldsymbol{x})$ is defined as:
\begin{equation}
\mathcal{L}(\boldsymbol{x}) = \mathbb{E}[l(y)] = \int l(y)p(y|\boldsymbol{x})dy
\end{equation}
This integral is often approximated using methods such as Monte Carlo estimation. Standard choices for acquisition functions include the expected improvement (EI) and the upper confidence bound~\cite{srinivas2009gaussian}.

\paragraph{Expected improvement.}  
The EI represents the average amount by which sampling a point $\boldsymbol{x}$ is expected to improve upon the current best-observed objective value. The improvement is defined by the utility function $l(y) = \max(0, y - f_{\text{max}})$. For a GP with predictive mean $\mu(\boldsymbol{x})$ and standard deviation $\sigma(\boldsymbol{x})$, the EI can be computed analytically as:
\begin{equation}
\label{eq:EI}
\mathcal{L}_{\text{EI}}(\boldsymbol{x}) = \mathbb{E}[\max(0, y - f_{\text{max}})] = \sigma(\boldsymbol{x}) \left[ Z \Phi(Z) + \phi(Z) \right]
\end{equation}
where $Z = \frac{\mu(\boldsymbol{x}) - f_{\text{max}}}{\sigma(\boldsymbol{x})}$, $\Phi(\cdot)$ is the cumulative distribution function of the standard normal distribution, and $\phi(\cdot)$ is the probability density function of the standard normal distribution.

\subsection{Cascade processes}\label{sec:cascade}
Many complex systems, in particular in materials science and chemistry, but also in optimisation problems beyond these disciplines, are structured as multistage processes, where a sequence of operations must be executed to achieve a final result. Examples are materials characterisation steps before thin-film property measurements or even device measurements; in-situ measurements of molecules taken during synthesis or after purification, followed by more complex measurements of molecular functionality; and generally any computations or simulations preceding experimental synthesis during molecular and materials design.

We define a cascade process as a specific type of multistage system presenting a sequential and dependent structure (see Fig.~\ref{fig:block_diag}). In a cascade process, the product, or output, ($\boldsymbol{h}_{i-1}$) of any given stage serves as input for the subsequent stage ($i$).
Formally, a cascade process consists of $N$ stages, where each stage $i$ is represented by a function $f_{i}$. The stage function depends on a set of controllable parameters $\boldsymbol{x}_{i}$ and the output from the preceding stage:
\begin{equation}
\boldsymbol{h}_{i} = f_{i}(\boldsymbol{x}_{i}, \boldsymbol{h}_{i-1}) + \boldsymbol{\epsilon}^{p}_i, \quad \text{for } i=1, \dots, N.
\end{equation}
where $\boldsymbol{\epsilon}^{p}_i$ is noise associated with the process.
The final system output is $\boldsymbol{y} = \boldsymbol{h}_{N}$. This structure is common in areas like chemical manufacturing, semiconductor fabrication, and material processing, where the outcome of an initial reaction or preparation step directly constrains and influences all subsequent steps. This dependency means that sub-optimal performance at an early stage propagates through the entire system, making the joint optimisation of all controllable parameters $\{\boldsymbol{x}_{i}\}_{i=1}^N$ a challenging problem.

%% file: 03_methods.tex
We present MSBO (multi-stage Bayesian optimisation), a Bayesian optimisation framework designed for sequential multi-stage processes. MSBO addresses the challenge of optimising complex chains where the output of each operation serves as the input for subsequent stages. The framework models the system's dependency structure by employing a cascade of GPs inspired by \citet{astudillo2021bayesian}, which is then paired with nested acquisition function evaluations to account for heterogeneous computational costs and parameter spaces.

\begin{figure}[H]
    \centering
    \includegraphics[width=0.9\columnwidth]{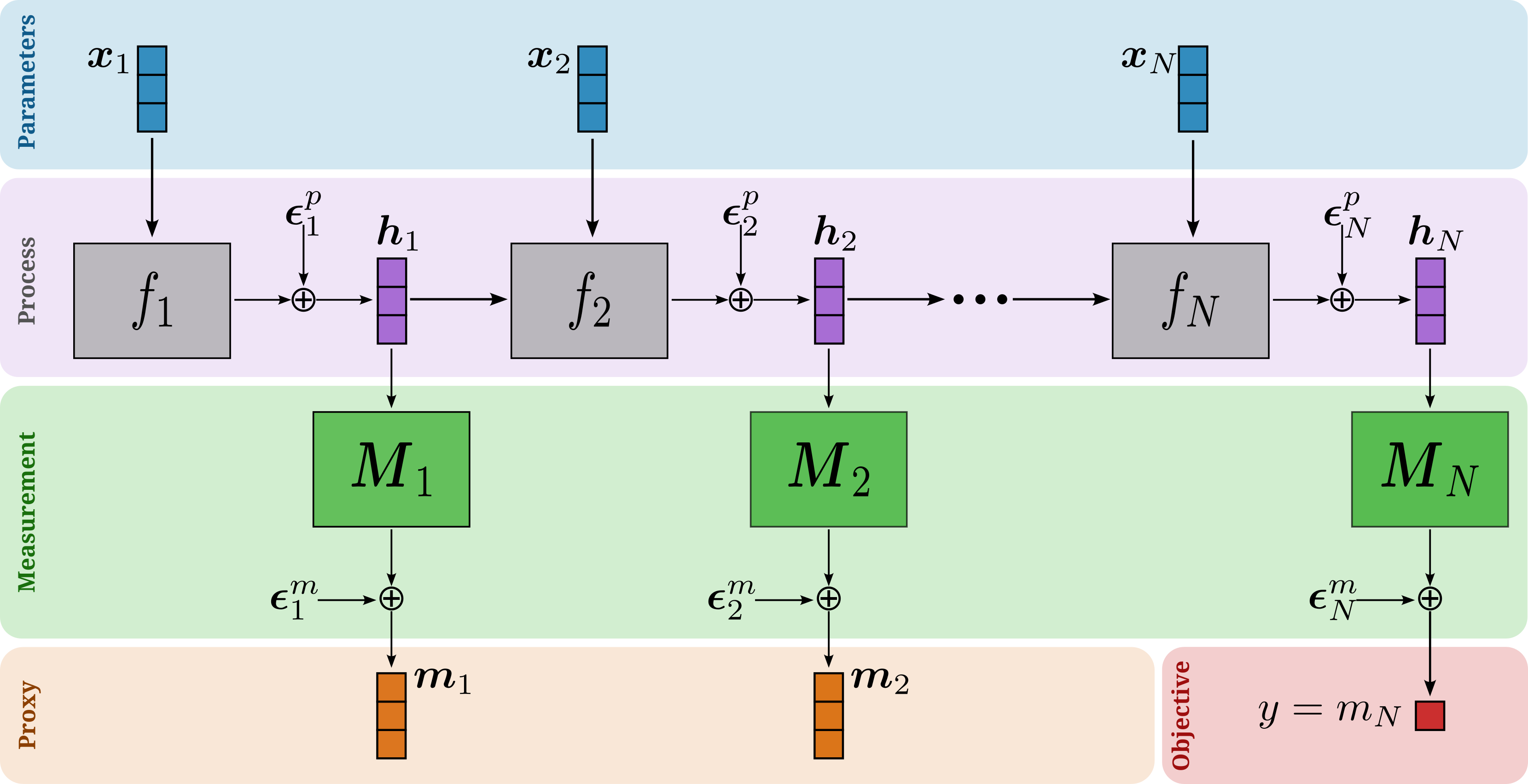}
    \caption{Illustration of a cascade process of N stages. At each stage i, the process function $f_i$ takes controllable parameters $\boldsymbol{x}_{i}$ and the latent output from the previous stage $\boldsymbol{h}_{i-1}$ to generate a new latent state $\boldsymbol{h}_{i}$, subject to process noise $\boldsymbol{\epsilon}^p_{i}$. The optimizer does not observe $\boldsymbol{h}_{i}$ directly but receives a proxy measurement $\boldsymbol{m}_{i}$, obtained via a masking operator $\boldsymbol{M}_i$ and additive measurement noise $\boldsymbol{\epsilon}^m_{i}$. The final objective is defined as the terminal measurement $y = m_N$.}
    \label{fig:block_diag}
\end{figure}

\paragraph{Optimisation objective.}
Consider a cascade process, as defined in Section~\ref{sec:cascade} and illustrated in Fig.~\ref{fig:block_diag}, consisting of $N$ distinct operations. Let $f_i$ denote the true (latent) function governing the $i$-th stage, where the intermediate output $\boldsymbol{h}_i$ is a function of the tunable inputs $\boldsymbol{x}_i$ and the preceding latent output $\boldsymbol{h}_{i-1}$. The key distinction for optimisation is that the true outputs $\boldsymbol{h}_i$ are not always directly observable. Instead, we obtain a measurement $\boldsymbol{m}_i$ of the intermediate output $\boldsymbol{h}_i$ at stage $i$, defined as:
\begin{equation}
\boldsymbol{m}_i = \boldsymbol{M}_i \boldsymbol{h}_i + \boldsymbol{\epsilon}^{m}_i
\end{equation}
where $\boldsymbol{M}_i$ is a (binary diagonal masking) matrix that selects a subset of the state $\boldsymbol{h}_i$ for observation (to model partially observable states), and $\boldsymbol{\epsilon}^{m}_i$ is the additive observation noise. A generalisation to arbitrary matrices $\boldsymbol{M}$ is straightforward, but not tested in this work. The overall objective is to maximise the final measured output $y = m_N$ through the optimal selection of the joint parameter set $\mathcal{X} = \{\boldsymbol{x}_1, \dots, \boldsymbol{x}_N\}$, where we assume the final measurement is scalar. While we restrict this study to scalar objectives, the framework can be readily extended to multi-objective BO.

\paragraph{Resumable sampling.}
A core feature of our framework is its capability to handle partially processed samples through the use of an inventory. Unlike conventional optimisation approaches that require the complete evaluation of the final objective function, MSBO permits stopping and resuming sample processing at any intermediate stage of the cascade. The inventory maintains a record of all sample states, process parameters, and measurements, tracking which sub-processes have been completed for each sample ID.

The use of an inventory allows for sample continuation: the optimisation can resume from any intermediate state, and partially processed sample states can be retrieved and applied as inputs to subsequent sub-processes. This system promotes data efficiency, since intermediate results are preserved and reused rather than being discarded, thereby avoiding redundant computations. Finally, this architecture supports dynamic resource allocation by allowing the algorithm to select which sub-process to sample next based on acquisition function values. 

This resumable sampling capability allows the algorithm to dynamically balance effort between advancing existing samples with high expected utility through subsequent evaluation stages versus initiating new samples to explore other regions of the parameter space. The resulting framework is particularly valuable in real-world scenarios, enabling dynamic decision-making and enhancing overall data efficiency.

\subsection{Cascade Gaussian processes}
\begin{figure}[H]
    \centering
    \includegraphics[width=0.9\columnwidth]{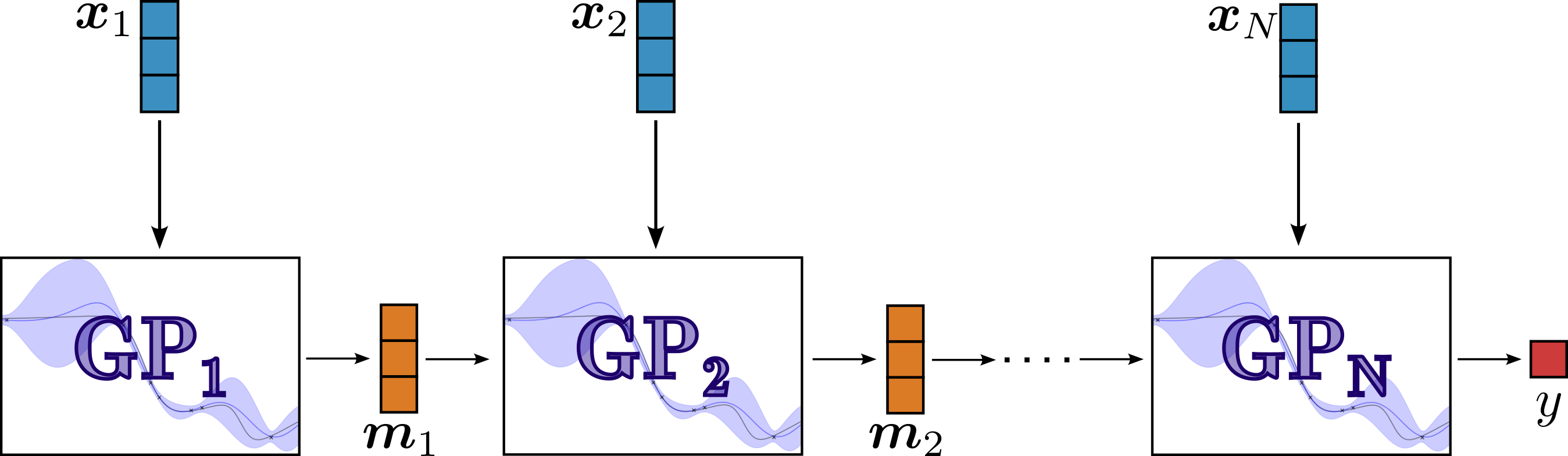}
    \caption{Schematic illustration of the standard cascade configuration of our surrogate model as a chain of independent GPs. The probabilistic output distribution of the upstream model $\text{GP}_{i-1}$ serves as a noisy input for the downstream model $\text{GP}_{i}$. Epistemic uncertainty is propagated through the chain, from the initial parameters $\boldsymbol{x}_1$ to the final objective $y$, via Monte Carlo sampling.}
    \label{fig:cascadeGP}
\end{figure}

To capture the sequential nature of the data, we model the system using a cascade of independent GPs (see Fig.~\ref{fig:cascadeGP}). Each stage $i$ is approximated by a probabilistic surrogate:
\begin{equation}
    f_i(\boldsymbol{x}_i^{\text{aug}}) \sim \mathcal{GP}\left(\mu_i(\boldsymbol{x}_i^{\text{aug}}), k_i(\boldsymbol{x}_i^{\text{aug}}, \boldsymbol{x}_i^{\prime\text{aug}})\right)
\end{equation}
where $\boldsymbol{x}_i^{\text{aug}}$ represents the augmented input vector. The framework supports two architectural configurations:
\begin{itemize}
    \item Standard cascade: $\boldsymbol{x}_i^{\text{aug}} = [\boldsymbol{x}_i, \boldsymbol{m}_{i-1}]$, creating a direct Markovian dependency.
    \item Residual connections: $\boldsymbol{x}_i^{\text{aug}} = [\boldsymbol{x}_i, \boldsymbol{x}_{i-1}, \boldsymbol{m}_{i-1}]$, enabling the model to preserve information from previous inputs, analogous to residual links in deep neural networks. This configuration helps mitigate information loss, especially with partially observed outputs.
\end{itemize}

To propagate the epistemic uncertainty (the uncertainty in the model due to the limited amount of training data) of the GPs along the chain, we implement a Monte Carlo sampling approach. For a cascade of $N$ processes, uncertainty propagation proceeds as follows: at stage $i=1$, we draw $S$ samples from the posterior of the first GP, $\boldsymbol{m}_1^{(s)} \sim \text{GP}_1(\boldsymbol{x}_1)$ for $s=1,\dots,S$. For subsequent stages $i>1$, each sample $\boldsymbol{m}_{i-1}^{(s)}$ from the previous process serves as an input to the next GP, yielding $\boldsymbol{m}_i^{(s)} \sim \text{GP}_k([\boldsymbol{x}_i, \boldsymbol{m}_{i-1}^{(s)}])$. This sampling-based approach naturally captures the non-Gaussian, potentially multimodal uncertainty distributions that emerge from the cascade of nonlinear transformations, providing realistic uncertainty quantification for the final process outputs. The resulting ensemble of samples $\{\boldsymbol{m}_K^{(s)}\}_{s=1}^S$ represents the compound uncertainty and enables acquisition functions to make decisions that account for uncertainty accumulation throughout the entire process cascade.

\subsection{Multi-stage expected improvement}
To fully leverage the resumable sampling capability described earlier, we propose a nested optimisation strategy. This approach constructs an acquisition function that takes into account the sequential dependency of the cascade process, where the utility of a decision at an early stage depends on the potential outcomes of subsequent stages.

We formulate the acquisition function recursively. For the final stage $N$, the acquisition function is the standard EI on the objective $y$, conditioned on the previous measurement $\boldsymbol{m}_{N-1}$. For any preceding stage $i < N$, the acquisition function is defined as the expected value of the acquisition function of the subsequent stage $i+1$, integrated over the probabilistic outcome of the current stage $\boldsymbol{m}_i$, thus allowing the uncertainty at intermediate stages to propagate forward to the final objective.

Formally, let $y^\ast$ denote the best objective value observed so far. The nested acquisition function $\alpha$ is defined recursively as:

\begin{equation}
    \label{eq:nested-ei}
    \left\{
    \begin{array}{l}
    \alpha_{\text{EI} \mid \boldsymbol{m}_{N-1}}(\boldsymbol{x}_N)
    = \mathbb{E}_{y \sim \mathcal{GP}_N(\boldsymbol{x}_N^{\text{aug}})}
        \left[ \max(0,\,y - y^\ast) \right] \\[1em]
    
    \alpha_{\text{EI} \mid \boldsymbol{m}_{N-2}}(\boldsymbol{x}_{N-1}, \boldsymbol{x}_N)
    = \mathbb{E}_{\boldsymbol{m}_{N-1} \sim \mathcal{GP}_{N-1}(\boldsymbol{x}_{N-1}^{\text{aug}})}
        \left[ \alpha_{\text{EI} \mid \boldsymbol{m}_{N-1}}(\boldsymbol{x}_N) \right] \\[1em]
    
    \qquad\qquad \vdots \qquad\qquad\qquad\qquad\qquad\qquad  \vdots \\[1em]
    
    \alpha_{\text{EI}}(\boldsymbol{x}_1, \ldots, \boldsymbol{x}_N)
    = \mathbb{E}_{\boldsymbol{m}_1 \sim \mathcal{GP}_1(\boldsymbol{x}_1)}
        \left[ \alpha_{\text{EI} \mid \boldsymbol{m}_1}(\boldsymbol{x}_2, \ldots, \boldsymbol{x}_N) \right]
    \end{array}
    \right.
\end{equation}

At each iteration, the optimiser queries the inventory to identify all valid continuation points, as well as the option to initialise a new sample at the first sub-process. For each candidate process $i$ and its corresponding available input $\boldsymbol{m}_{i-1}$ (retrieved from the inventory), we optimise the parameters $\boldsymbol{x}_i$ to maximise the nested acquisition function. To prevent stagnation in regions where the EI vanishes (i.e., is approximately zero across the search space), the algorithm automatically switches to the upper confidence bound acquisition function with a high exploration parameter.

The final selection of the next experiment is determined by comparing the maximal acquisition values across all stages. The framework allows for cost-weighted selection to account for heterogeneous experimental costs $c_i$ per stage:

\begin{equation}
    \text{next stage} = \operatorname*{argmax}_{i, \boldsymbol{x}_i} \frac{\alpha(\dots)}{c_i}.
\end{equation}

We found in our experiments that setting a uniform cost ($c_i = 1$) for all stages generally leads to superior optimisation performance compared to explicitly modelling the cost ratios. This approach ensures that the algorithm prioritises information gain and the potential for high-performing solutions over short-term cost savings. 

\subsection{Sampling strategy}
The experimental design strategy consists of two phases: First, the acquisition of an initial dataset $\mathcal{D}_0$ through quasi-random sampling of the entire cascade to build an initial surrogate model. To this aim, we use Sobol sequences, ensuring space-filling properties in the high-dimensional joint input space. In all experiments, the size of $\mathcal{D}_0$ was set to $2(D+1)$, where $D$ is the total number of controllable parameters (components of all $\boldsymbol{x}_i$ for $i = 1, \ldots, N$). This is followed by an adaptive optimisation phase, where new sampling points are selected based on the acquisition function. In this adaptive phase, the algorithm can choose from which sub-process to sample next, rather than being restricted to evaluating the entire cascade from the initial input. We allow users to define a constraint on the minimal relative frequency at which each sub-process is sampled. This constraint is essential in settings where the final sub-process is relatively easy to model. In this scenario, the high uncertainty propagating from complex upstream stages can dominate the acquisition function, leading the algorithm to undersample the final stage and consequently fail to realise the optimal set of parameters (i.e., the best objective value).

\subsection{Implementation details}
All GPs use scaled radial basis function kernels with automatic relevance determination. All inputs are normalised to $[0,1]^d$, and outputs are standardised to zero mean and unit variance. Hyperparameters are optimised via marginal likelihood maximisation at each iteration. The acquisition function optimisation employs a multi-start gradient-based approach (L-BFGS-B) with Sobol initialisation to ensure global search capability. For experiments involving discrete input spaces, the acquisition function is instead computed over all possible inputs, and the maximum is selected via exhaustive search.

%% file: 04_experiments.tex
To validate our MSBO framework, we perform a comprehensive evaluation across synthetic functions and real-world molecular property prediction tasks. We demonstrate the superior data efficiency and robustness of MSBO, arising from its ability to explicitly model cascade processes and dynamically manage intermediate measurements, compared to standard methodologies. We ablate the contributions that stem from the cascade GP and the flexible multi-stage acquisition function with the resumable sampling mechanism.

\subsection{Experimental setup}
\label{exp_setup}

\paragraph{Baselines.}
We compare MSBO against three baselines:
\begin{itemize}
    \item Random search: In each iteration, the set of candidate parameters $\boldsymbol{x} = (\boldsymbol{x}_1, \dots, \boldsymbol{x}_N)$ is drawn from a uniform distribution. This establishes a lower bound for optimisation performance.
    \item Standard BO: This baseline treats the entire multi-stage process as a single black-box function, optimising the final objective $y(\boldsymbol{x}_1, \dots, \boldsymbol{x}_N)$ without leveraging intermediate measurements. It uses the expected improvement acquisition function.
    \item BOFN (Bayesian optimisation of function networks) \cite{astudillo2021bayesian}: This baseline uses the known function network structure and intermediate measurements through our cascade GP as a surrogate model, but does not implement resumable sampling or inventory management. It uses the expected improvement acquisition function.
\end{itemize}

\paragraph{Metrics.}
Our primary metric for evaluating performance is the logarithm of the simple regret,
\[
R_t = \log(\left| y_{\text{opt}} - y_t^\ast \right|),
\]
where $y_{\text{opt}}$ is the global optimum of the objective function and $y_t^\ast$ is the best objective value observed up to iteration $t$. In settings with process noise, noisy samples may occasionally yield values exceeding the true, denoised optimum. To address this, we define the reference $y_{\text{opt}}$ as the true optimum plus three times the standard deviation of the cumulative output noise. This prevents the regret metric from becoming undefined or artificially saturated due to stochastic outliers. For the real-world task, we report percentile-based measures that reflect the best-performing fraction of the dataset discovered by the optimiser.

\paragraph{Heterogeneous costs.}
Unless otherwise specified, we adopt a uniform cost model in which generating a sample for any sub-process (i.e., completing any stage $i$) consumes $c_i=1$ unit of budget. This allows a direct comparison based on sampling efficiency. We also consider heterogeneous cost scenarios to simulate more expensive downstream stages (Figs.~\ref{fig:sweep_cost}, \ref{fig:percentiles_summary}). In those experiments, we normalise the total cost of sampling the whole cascade to $1$ so that each full evaluation consumes the same total budget, but the cost is fractioned differently between sub-processes; this normalisation simplifies visual comparison of optimisation trajectories under different cost repartitions.

\subsection{Synthetic benchmarks}

We benchmark MSBO using our synthetic function generator (details in SI~\ref{app:synthetic}), which provides fine-grained control over function complexity, input dimensionality, and noise properties, thereby creating a controlled environment for evaluating our framework. The generator constructs multi-stage processes by chaining interconnected neural networks, where each stage is trained on a random seed dataset to create continuous, differentiable functions. By varying the size of these seed datasets, we can implicitly tune the complexity of each sub-process while simulating realistic constraints such as information masking and error propagation.

\paragraph{Conceptual demonstration of MSBO sampling efficiency.}
We first examine a simple two-stage cascade $f_1(x_1) \rightarrow f_2(h_1, x_2) \rightarrow y$,
with both $x_1\text{ and }x_2 \in \mathbb{R}$, so that both landscapes can be visualised directly. In these experiments, we set $\boldsymbol{M}=\boldsymbol{I}$ and both process and measurement noise to $0$. The complexity of the first and second stages is set to 8 and 2, respectively. In our framework, complexity is controlled by the number of random seed points used to construct the synthetic functions, where higher values correspond to more rugged landscapes and increased modelling difficulty of the objective (see SI~\ref{app:synthetic}).

\begin{figure}[H]
    \centering
    \includegraphics[width=0.9\columnwidth]{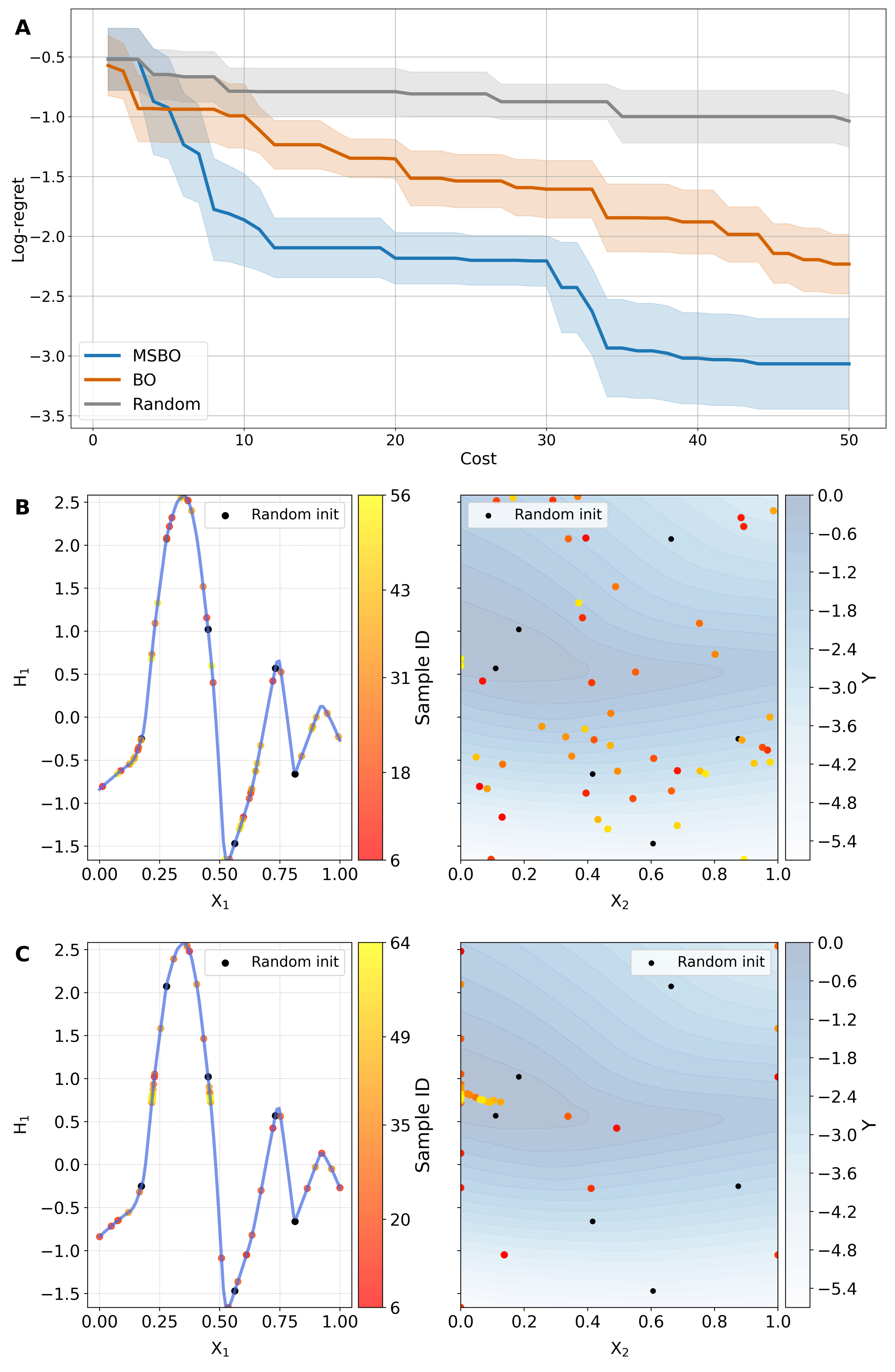}
    \caption{MSBO achieves superior data efficiency by selectively allocating samples to promising regions of the parameter space. (A) Aggregated log-regret over 10 independent runs as a function of sampling cost (uniform cost model), with shaded regions indicating single standard deviation interval on both sides of the mean. Both cost and regret are unitless. (B) and (C) show a representative sampling distribution for standard BO (B) and MSBO (C), respectively. The left subplots display the first-stage intermediate outputs $h_{1}$ (dots) plotted against the input $x_{1}$, overlaid on the true first-stage function (blue line). The right subplot shows the locations of the second-stage evaluations within the $h_{1} \times x_{2}$ domain. Markers in both plots are colour-coded by the sample ID (iteration number), with black dots representing the samples of the random initialisation.}
    \label{fig:small_process}
\end{figure}

We find that, for any given cost, MSBO consistently achieves lower regret than the standard BO baseline and random search, demonstrating improved data efficiency (see Fig.~\ref{fig:small_process} for representative results from these experiments). All experiments are averaged over 10 independent runs to obtain uncertainty bounds and to ensure statistical significance. Furthermore, while MSBO suggests a similar number of first-stage function evaluations as BO (left panels in Fig.~\ref{fig:small_process}B and C), it allocates second-stage resources more efficiently, concentrating samples in the region of the global optimum. In the standard BO run, the samples for the second process are broadly scattered across the domain and are not concentrated near the true optimum of the second-stage landscape. By contrast, MSBO concentrates the majority of its second-stage samples in the region close to the maximum of the second-stage objective, showing that the method is substantially more efficient at allocating second-stage evaluations to informative regions.

MSBO discovers and explores multiple high-performing modes in the first process: samples are clustered around several distinct regions of the $f_1$ landscape that correspond to high-valued objectives, indicating the algorithm's ability to identify and maintain promising alternatives rather than collapsing prematurely to a single local mode. This behaviour is a direct consequence of (i) the cascade GP that propagates multi-modal uncertainty, and (ii) the nested, resumable acquisition strategy that allows the optimiser to selectively invest further budget only for the most promising samples, pruning the search space early. More examples are provided in SI~\ref{app:additional} (Figs.~\ref{fig:qualitative1}-\ref{fig:qualitative4}).

\paragraph{Complexity sweep.}
We now assess the behaviour of MSBO on more challenging synthetic tasks where the two-stage cascade process is defined as $f_1(\boldsymbol{x}_1) \rightarrow f_2(\boldsymbol{h}_1, \boldsymbol{x}_2) \rightarrow y$. In this setting, the control parameters are $\boldsymbol{x}_1 \in \mathbb{R}^4$, the intermediate observables are $\boldsymbol{h}_1 \in \mathbb{R}^2$, and the second-stage control parameters are $\boldsymbol{x}_2 \in \mathbb{R}^2$. We assume fully observable states ($\boldsymbol{M}=\boldsymbol{I}$) and no noise. The synthetic generator allows us to independently adjust the complexity of both the first ($f_1$) and second ($f_2$) sub-processes (see SI~\ref{app:synthetic} for details). This enables a broad exploration across different relative complexity settings.

\begin{figure}[H]
    \centering
    \includegraphics[width=0.95\columnwidth]{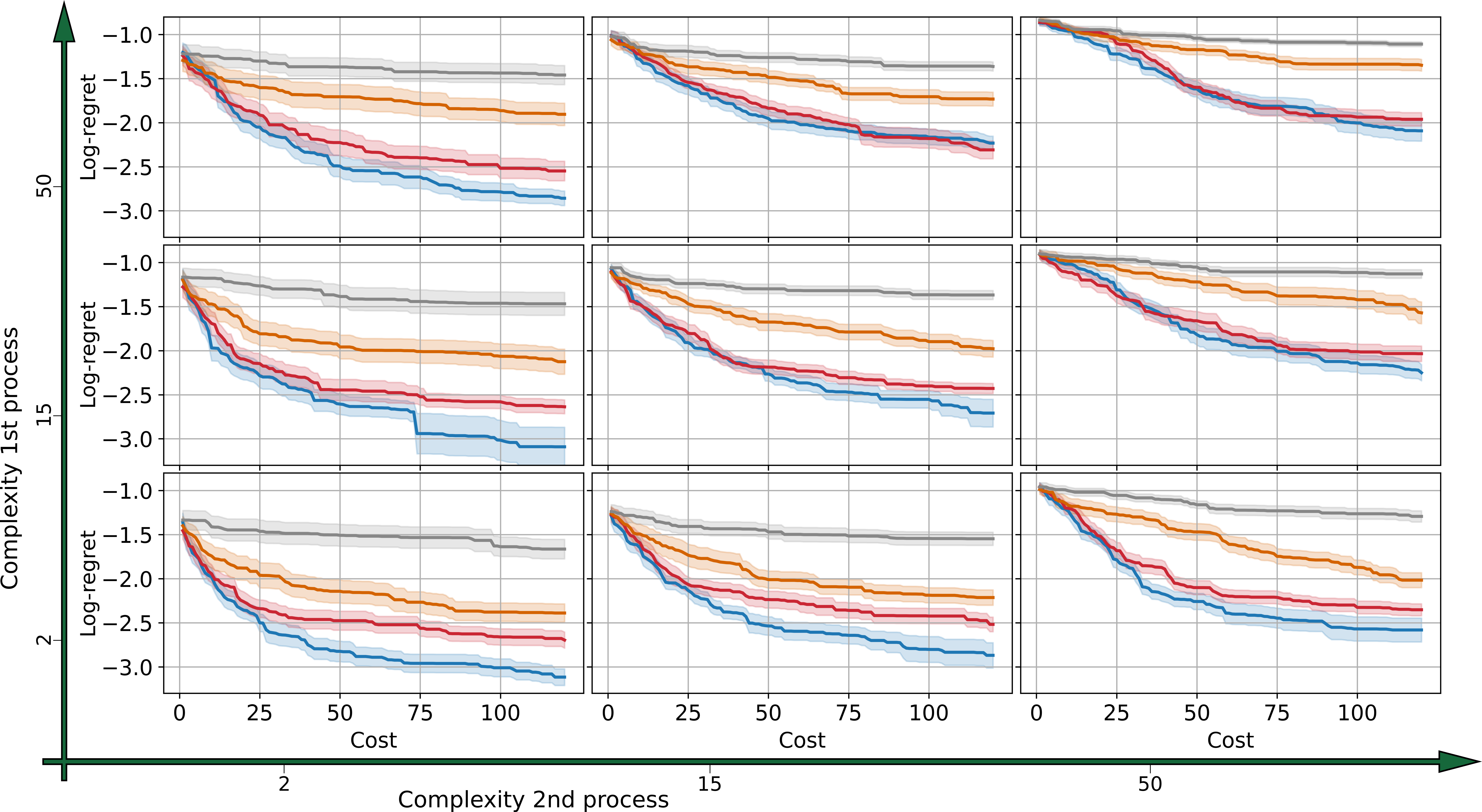}
    \caption{Log-regret plotted against normalised sampling cost (uniform cost model) for a two-stage cascade process. All results are aggregated over 20 optimisation runs. Shaded regions indicate a 1 standard deviation interval. The rows correspond to increasing complexity levels for the first stage ($\mathcal{D}_{\text{seed}}$ sizes 2, 15, 50). The columns correspond to increasing complexity levels for the second stage ($\mathcal{D}_{\text{seed}}$ sizes 2, 15, 50). Each panel contains performance curves for MSBO (blue), standard BO (orange), BOFN (red), and random search (grey). Cost and regret are considered unitless here.}
    \label{fig:sweep_regret}
\end{figure}

To explore different possible real-world scenarios, our experiments test nine combinations of first- and second-stage complexities. We compare MSBO with standard BO (which treats the cascade as a monolithic black box) and the BOFN baseline, which can also use the intermediate measurement $\boldsymbol{m}_1=\boldsymbol{h}_1$. We find that MSBO consistently outperforms the other baselines in all scenarios. The ability of MSBO and BOFN to learn the intermediate mapping $\hat{h}_1(\boldsymbol{x}_1)$ provides a strong driving signal, overcoming the difficulties standard BO faces in optimising over the high-dimensional joint space $(\boldsymbol{x}_1, \boldsymbol{x}_2)$. This is clearly demonstrated by the significant performance gap that both MSBO and BOFN exhibit with respect to BO across the entire sweep. Furthermore, MSBO's multi-stage acquisition function and support for resumable sampling provide a substantial, ulterior boost in performance over all baselines, including BOFN. MSBO consistently achieves the lowest regret for a given cost and maintains a clear margin over the baselines. This advantage persists whether the difficulty arises primarily in the first stage (left column), in the second stage (bottom row), or simultaneously in both. The results are illustrated in Fig.~\ref{fig:sweep_regret}, which reports the mean simple regret as a function of normalised sampling cost, aggregated over $20$ independent runs. 

Specifically, our results suggest that MSBO’s advantage is not confined to specific complexity patterns: the ability to decouple upstream and downstream exploration, and to commit second-stage resources only when justified by intermediate observations, yields systematic improvements. This is evident not only when the initial process ($f_1$) is more complex (top-left corner), but crucially, even when the second process ($f_2$) is the most difficult to regress (bottom-right corner). This observation suggests substantial opportunities and potential for applying MSBO in scenarios where no highly informative proxy measurement is available. Even in such cases, low-fidelity and not directly related observations, additional experiments, as well as simulations can improve the overall sample efficiency of complex and costly experiments of the actual objective function.

Furthermore, we observe that the relative sampling frequency of first-stage and second-stage evaluations adjusts dynamically, demonstrating the framework's ability to allocate resources depending on the relative complexity of the stages. We report the number of samples drawn from each of the two subprocesses for each setting in SI~\ref{app:additional}, Figure~\ref{fig:count_hist}.

\paragraph{Impact of heterogeneous costs.}
A common scenario in experimental campaigns is that downstream characterisation, such as device fabrication, purification, or final testing, is substantially more expensive than upstream preparation or proxy stages, which often involve inexpensive computations or early-stage synthesis.

\begin{figure}[H]
    \centering
    \includegraphics[width=0.75\columnwidth]{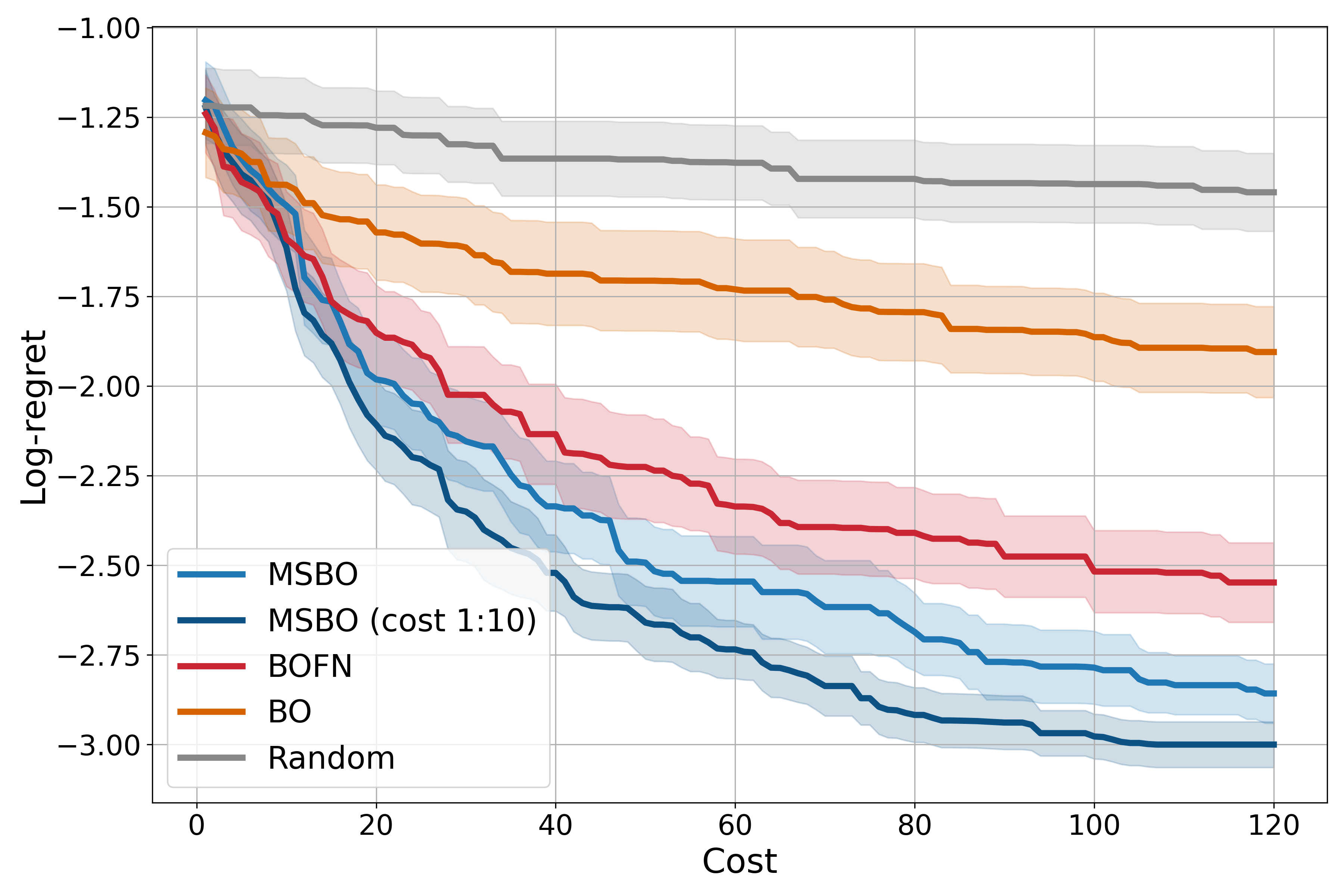}
    \caption{Log-regret versus cumulative cost for a two-stage process with first- and second-stage complexities of 50 and 2, respectively. The light blue curve represents MSBO in the standard uniform cost setting (ratio 1:1), while the dark blue curve represents the heterogeneous cost setting where the second stage is ten times more expensive than the first (ratio 1:10). The baselines always execute the full cascade sequence; as the total cost is normalized to 1, their trajectories are identical in both settings. The trajectories represent the aggregated results over 20 optimisation runs. Shaded regions indicate a 1 standard deviation interval. Cost and regret are considered unitless here.}
    \label{fig:sweep_cost}
\end{figure}

To evaluate how robust MSBO remains under such realistic conditions, we examine its behaviour when the relative costs of the two stages are varied. We focus on the process defined by complex upstream dynamics (corresponding to the top-left case in Fig.~\ref{fig:sweep_regret}) and compare two regimes: a uniform-cost setting (ratio 1:1) and a heterogeneous-cost setting (ratio 1:10), where the second stage is ten times more expensive than the first (see Fig.~\ref{fig:sweep_cost}). As detailed in Section~\ref{exp_setup}, the total cost of a full cascade evaluation is normalised to 1 in all experiments. Consequently, the baseline methods (standard BO and BOFN), which do not support dynamic early stopping, exhibit identical behaviour regardless of the internal cost distribution. By contrast, MSBO naturally adapts to the cost asymmetry through its resumable sampling mechanism. Even though MSBO does not explicitly incorporate cost parameters into the acquisition function, its resumable sampling mechanism allows it to implicitly capitalise on low-cost intermediate evaluations.

Empirically, MSBO exhibits improved data efficiency in the heterogeneous-cost regime compared to the uniform one. The MSBO algorithm relies more heavily on the inexpensive first stage to discard unpromising candidates early, and only progresses samples to the costly second stage when their intermediate results indicate sufficiently high utility. This behaviour leads to faster convergence toward better-performing solutions.

\paragraph{Three-stage experiments and noisy observations.}
We conduct further experiments to assess the framework's scalability to longer cascades and its robustness in more realistic settings, specifically addressing process noise (imperfectly reproducible experiments) and partial observability of sample states ($\boldsymbol{M} \neq \boldsymbol{I}$).

When evaluating our model on a three-stage cascade process, we observe that the performance advantages of MSBO, as demonstrated in the two-stage setting, extend effectively to this deeper process, where it consistently outperforms the baselines (see aggregated results in Fig.~\ref{fig:three_steps}). To test the impact of partial observability of intermediate results, we evaluate our model in a more complex three-stage cascade, where each intermediate process outputs a four-dimensional state, of which only the first two dimensions are accessible to the optimiser. Consequently, half of the information required to fully model each stage is missing. Due to the increased complexity and the information loss, both standard BO and BOFN degrade to the performance level of random search, whereas MSBO remains significantly more efficient (see Fig.~\ref{fig:three_steps_filtered}). This result is significant, as it reflects the more realistic experimental setting, where not the entire intermediate result of an experiment is observable, but only aspects of it, e.g. a thin film material is generated, but only specific aspects of it are observed (e.g. conductivity or absorption), but not all details (microstructure, interfaces, other thin-film properties, etc.).

Finally, to assess robustness in realistic laboratory settings, we evaluate performance on a two-stage cascade with process noise $\boldsymbol{\epsilon}_{1}^p$. MSBO demonstrates superior robustness in this challenging regime compared to the baselines, consistently identifying high-performing solutions by effectively leveraging intermediate measurements and flexibly sampling from the various subprocesses.
Furthermore, calculating the denoised performance of the sample with the best observed objective $y$ reveals that relying on single observations often leads to selecting suboptimal parameters driven by favourable noise outliers.
Conversely, selecting the final parameters based on the surrogate model's maximum predicted mean among the collected samples consistently yields superior performance. This result suggests that in noisy real-world environments, a model-derived policy for selecting the final best-performing set of parameters is significantly more reliable than one based strictly on historical observations, as the model effectively filters out stochastic fluctuations. This advantage is particularly pronounced for MSBO. Because MSBO models the underlying process structure more accurately than the baselines, the performance gain achieved by relying on its surrogate, rather than on noisy observations, is significantly larger (see Fig.~\ref{fig:noisy}).

\subsection{Real-world applications}

We evaluate MSBO on three datasets representative of chemical optimisation tasks, comprising both computational and experimental workflows. These case studies exemplify the relevance of our method for real-world discovery campaigns, where multi-stage processes enable cost-efficient navigation of, for example, high-dimensional chemical spaces by bridging the gap between lower-fidelity proxies and expensive validations. In all scenarios, the objective is the optimisation of specific molecular properties. To manage the high dimensionality of the chemical space, we generate Morgan fingerprints (1024 dimensions, depth 3) for all molecules using RDKit~\cite{greg_landrum_2024_10633624} and compress these representations to the 16 most informative principal components using PCA.

\paragraph{QM9 (HOMO and LUMO levels).}
We target the minimisation of the energy levels of the highest occupied molecular orbital (HOMO) and the lowest unoccupied molecular orbital (LUMO) using the QM9 dataset \cite{ramakrishnan2014quantum}. These energy levels are critical indicators of optoelectronic behaviour and chemical reactivity. The property values in the dataset are derived from a high-fidelity, two-stage computational approach \cite{fediai2023accurate, fediai2023interpretable}. This task naturally aligns with a multi-stage workflow: the initial stage employs a density functional theory calculation using the Perdew-Burke-Ernzerhof (PBE) exchange-correlation functional with an aug-cc-TZVP basis set. This acts as a lower-fidelity stage-1 observation required for the second, high-fidelity stage, which involves refining the result using an accurate, but computationally expensive, eigenvalue-self-consistent GW approach (ev-GW@PBE/(aug-cc-DZVP$\to$aug-cc-TZVP)) \cite{ruddigkeit2012enumeration}. We estimate the computational cost of the initial DFT simulation to be approximately $1/10$ of the expensive GW correction.

\begin{figure}[H]
    \centering
    \includegraphics[width=0.9\columnwidth]{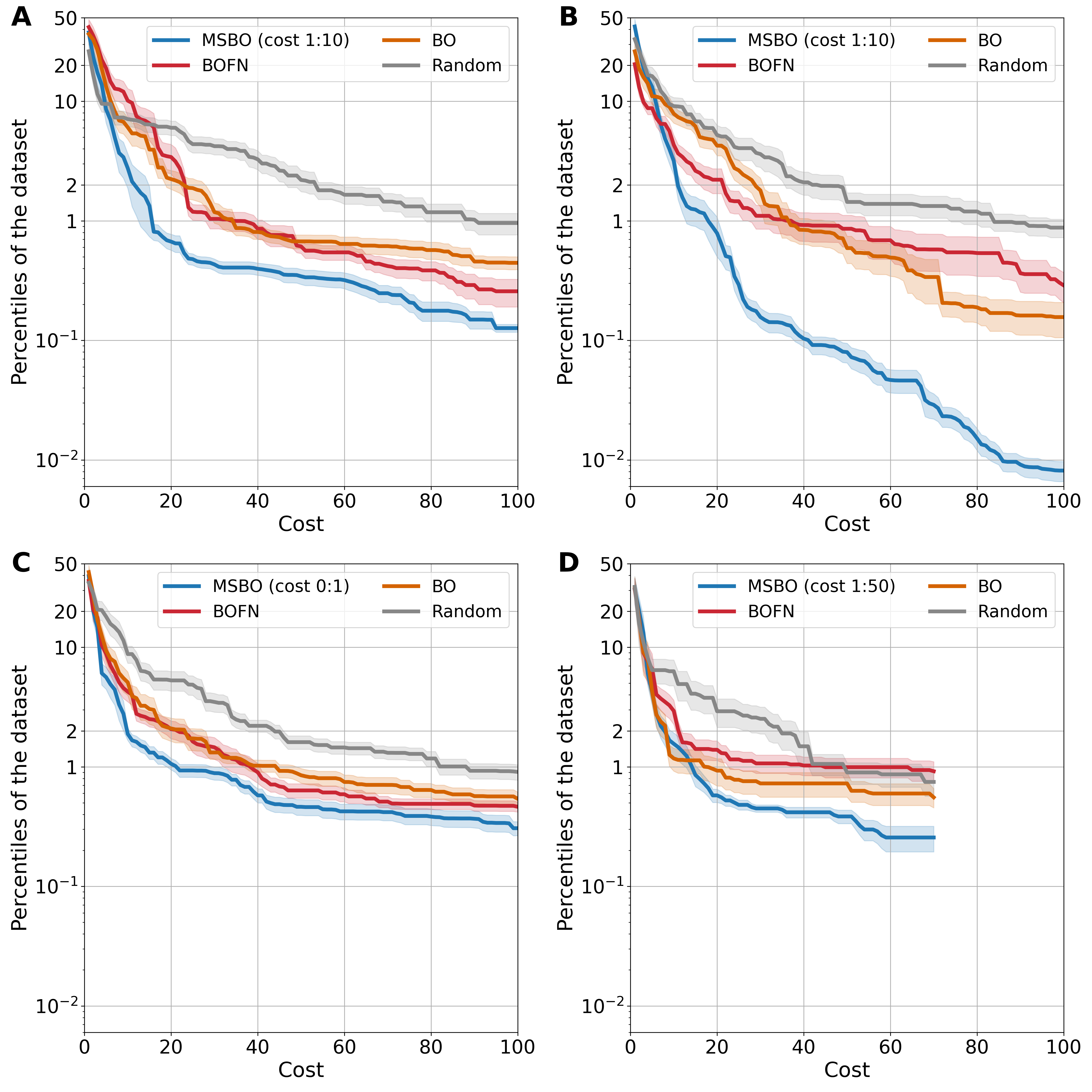}
    \caption{Percentile of the best-found molecule relative to the total dataset size as a function of accumulated cost. All results are aggregated over 30 optimisation runs. Shaded regions indicate a 1 standard deviation interval. The properties optimised, and the relative datasets are: (A) HOMO energy levels (QM9); (B) LUMO energy levels (QM9); (C) aqueous solubility (AqSolDB); and (D) hydration free energy (FreeSolv).}
    \label{fig:percentiles_summary}
\end{figure}

\paragraph{AqSolDB (solubility).}
For the minimisation of aqueous solubility, we utilise the AqSolDB dataset \cite{sorkun2019aqsoldb}, a large-scale, high-quality compilation focused on aqueous solubility values ($\log S$) for nearly 10,000 unique chemical compounds. The database aggregates and curates data from nine different public datasets and includes corresponding molecular representations. In our optimisation setup, we employ the partition coefficient (logP), computed via RDKit, as a proxy for the experimental solubility values. Given that the computational cost of calculating logP is negligible compared to the cost of performing wet-lab experiments, we treat the first-stage cost as effectively zero. In this setting, we use the residual connection configuration for our surrogate model (as described in Section~\ref{sec:cascade}) to mitigate the information loss inherent in compressing the molecular features into a single scalar proxy.

\paragraph{FreeSolv (hydration energy).}
We minimise the hydration free energy using the FreeSolv database \cite{mobley2014freesolv}, which contains values for the aqueous free energy of solvation ($\Delta G_{\text{solv}}$) for small, drug-like molecules. This property quantifies the change in free energy when a molecule moves from the gas phase to an aqueous solution. The database compiles two types of data: experimental values derived from a literature survey of reliable, peer-reviewed sources, and calculated values derived from alchemical free energy calculations performed using molecular dynamics simulations. The computational values were generated using the GAFF small molecule force field in TIP3P water with AM1-BCC charges, with simulations executed using the GROMACS package. In our experiments, the objective is to optimise the experimental value, using the values derived from MD simulations as the intermediate proxy. We estimate the cost of the simulation to be $1/50$ of the cost of the physical experiment.

\paragraph{Results.}
MSBO consistently outperforms the baselines, identifying molecules within the top 1\% of the dataset significantly faster. We quantify this performance across all four optimisation tasks by tracking the percentiles of the best-found solutions relative to the total dataset size as a function of accumulated cost (see Fig.~\ref{fig:percentiles_summary}). Notably, for all tasks excluding FreeSolv, MSBO discovers molecules in the top 1\% using only half of the budget required by standard BO. Moreover, the advantage in cost grows larger for lower percentiles, and MSBO is the only method that consistently identifies samples within the top 0.3\% in all datasets, within the allocated budget. This advantage is most dramatic in the specific case of the LUMO property, where MSBO navigates to the top 0.01\% of candidates, an order of magnitude lower than standard BO (see Panel B).

Furthermore, the comparison with BOFN reveals that simply modelling the cascade structure is insufficient. In the FreeSolv task, BOFN performs worse than standard BO, whereas MSBO's resumable sampling strategy effectively leverages the proxy measurements to outperform the baselines (see Panel D). Finally, MSBO exhibits the most rapid optimisation trajectory in the early stages (costs 0-20), indicating superior initial exploration and immediate identification of high-utility regions. These results confirm that the cascade structure provides a robust navigational signal, enabling efficient optimisation even with high-dimensional molecular representations and varying regimes of proxy fidelity and cost.

%% file: 05_conclusion.tex
In this work, we introduced MSBO, a Bayesian optimisation framework designed to enhance dynamic decision-making in self-driving laboratories (or multi-step computational workflows) with multi-stage cascade processes. Unlike standard Bayesian optimisation, which treats experimental workflows as rigid, monolithic black boxes, MSBO explicitly models the structure of cascade processes. By integrating a cascade GP surrogate with a nested acquisition function and an inventory management system, our approach enables resumable sampling, allowing the optimiser to selectively advance promising candidates while discarding uninformative ones at early stages.

We validated the framework through extensive synthetic benchmarking, demonstrating that MSBO consistently achieves lower regret than standard baselines and network-based alternatives (BOFN), and using a smaller budget. These experiments confirmed the method's robustness across various regimes of stage complexity and heterogeneous cost ratios. Furthermore, our evaluation of real-world optimisation tasks, including the optimisation of molecular solubility and electronic properties, revealed significant efficiency gains. Notably, MSBO identified top-performing molecules with considerably fewer resources than traditional methods, validating its practical utility in high-dimensional search spaces.

Ultimately, these results suggest that incorporating intermediate proxy measurements is a critical step toward more autonomous and efficient materials discovery. By bridging the gap between cheap computational approximations and expensive experimental validations, MSBO offers a scalable pathway for hybrid computational-experimental self-driving labs. The MSBO approach can replace (potentially biased) multi-stage funnel approaches, where selection criteria need to be defined beforehand, and can thus efficiently learn to optimise in high-dimensional parameter spaces with multiple intermediate observations. This opens possibilities to extend self-driving labs from confined optimisation tasks to more complex open challenges of autonomous discovery.

%% file: 07_code.tex
The code and supporting data are available upon reasonable request from the corresponding author, while a clean repository is being prepared for public release.

%% file: 10_acknowledgements.tex


We acknowledge support by the Federal Ministry of Education and Research (BMBF) under Grant No. 01DM21002A (FLAIM). 
This work was performed on the HoreKa supercomputer funded by the Ministry of Science, Research and the Arts Baden-Württemberg and by the Federal Ministry of Education and Research.



%% file: appendix.tex
\section{Synthetic dataset generator} \label{app:synthetic}

This section details the dataset generator we developed to benchmark sequential optimisation algorithms by creating complex, differentiable, multi-step functions. The problems are constructed using interconnected neural networks, offering tunable complexity, realistic constraints (e.g., additive measurement noise, additive process noise, information masking), and full reproducibility. A key feature is that the true global optimum is accessible via gradient-based methods, which facilitates the comparison and validation of different optimisation algorithms.

\subsection{Random function generation}

To establish a comprehensive testing framework for optimisation algorithms targeting cascade processes, we developed a system that generates continuous and differentiable random functions using neural networks. These functions are subsequently composed to form interdependent, multi-step cascades. Each constituent function is implicitly defined by training a multi-layer perceptron on a randomly sampled seed dataset, $\mathcal{D}_{\text{seed}} = (\boldsymbol{X}_{\text{seed}}, \boldsymbol{Y}_{\text{seed}})$.

\paragraph{Dataset and boundary constraints.}
The input data, $\boldsymbol{X}_{\text{seed}}$, are generated using Sobol sequences to ensure a quasi-random, uniform coverage within the $d$-dimensional hypercube $[0, 1]^d$. The corresponding target outputs, $\boldsymbol{Y}_{\text{seed}}$, are independently drawn from a standard normal distribution, $\mathcal{N}(0, 1)$. To prevent the resulting function from exhibiting maxima at the domain boundaries, we employ a mechanism for enforcing an artificial boundary penalty. Points sampled slightly outside the $[0, 1]^d$ domain are assigned a low target value (e.g., $-1.5$). This mechanism effectively pushes the function's response down near the edges, thereby encouraging the presence of internal maxima.

\begin{figure}[H]
    \centering
    \includegraphics[width=0.95\columnwidth]{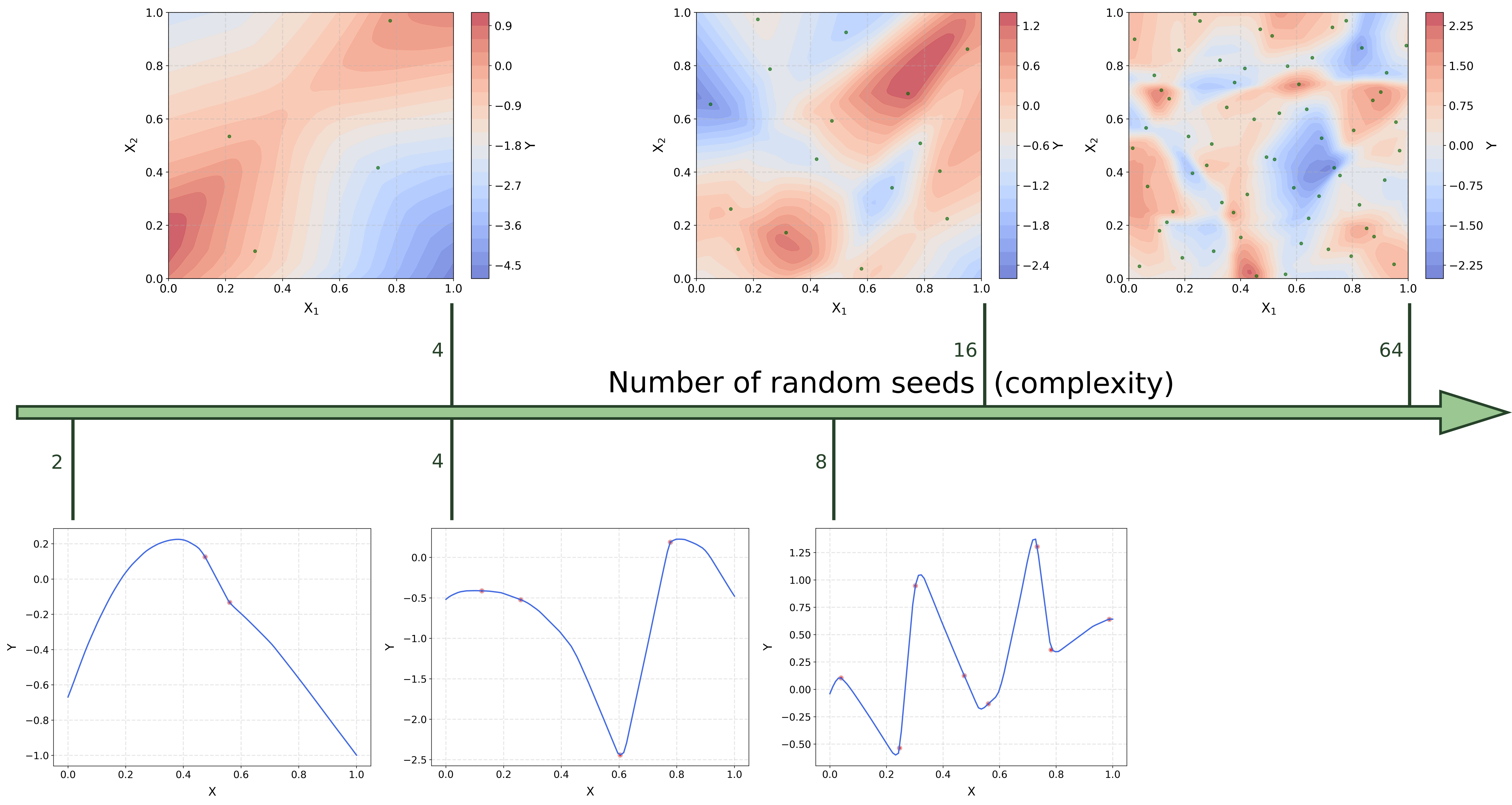}
    \caption{Examples of synthetic functions generated with varying sizes of the seed dataset $\mathcal{D}_{\text{seed}}$. The bottom row shows $1D$ functions yielding scalar outputs, while the top row shows functions in a $2D$ input space yielding scalar outputs. The size of $\mathcal{D}_{\text{seed}}$ directly controls the complexity of the resulting function landscape.}
    \label{fig:examples_generator}
\end{figure}

\paragraph{Neural network architecture and training.}
The function $f$ is implemented as a multi-layer perceptron with a customizable number of hidden layers and configurable dimensions. In our experiments, we use three hidden layers with dimensions $[64, 128, 32]$ and LeakyReLU as the activation function for all hidden layers. The network is trained for 800 epochs using the Adam optimiser ($\text{learning rate} = 0.01$) to minimise the mean squared error between the network's predictions and the random target data, using a batch size of 16. The final output of the network, $\boldsymbol{h}$, is optionally scaled to restrict its range using the sigmoid function. In our multi-step experiments, this scaling is applied to the output of all sub-processes except the final step, thus constraining the input of subsequent steps to the range $[0, 1]$.

Figure \ref{fig:examples_generator} illustrates examples of the generated functions in a 1D and a 2D parameter space for different sizes of the seed dataset $(\mathcal{D}_{\text{seed}})$. The size of this randomly sampled seed dataset directly correlates with what we define as the \textit{complexity} of the generated function.

\begin{figure}[H]
    \centering
    \includegraphics[width=0.95\columnwidth]{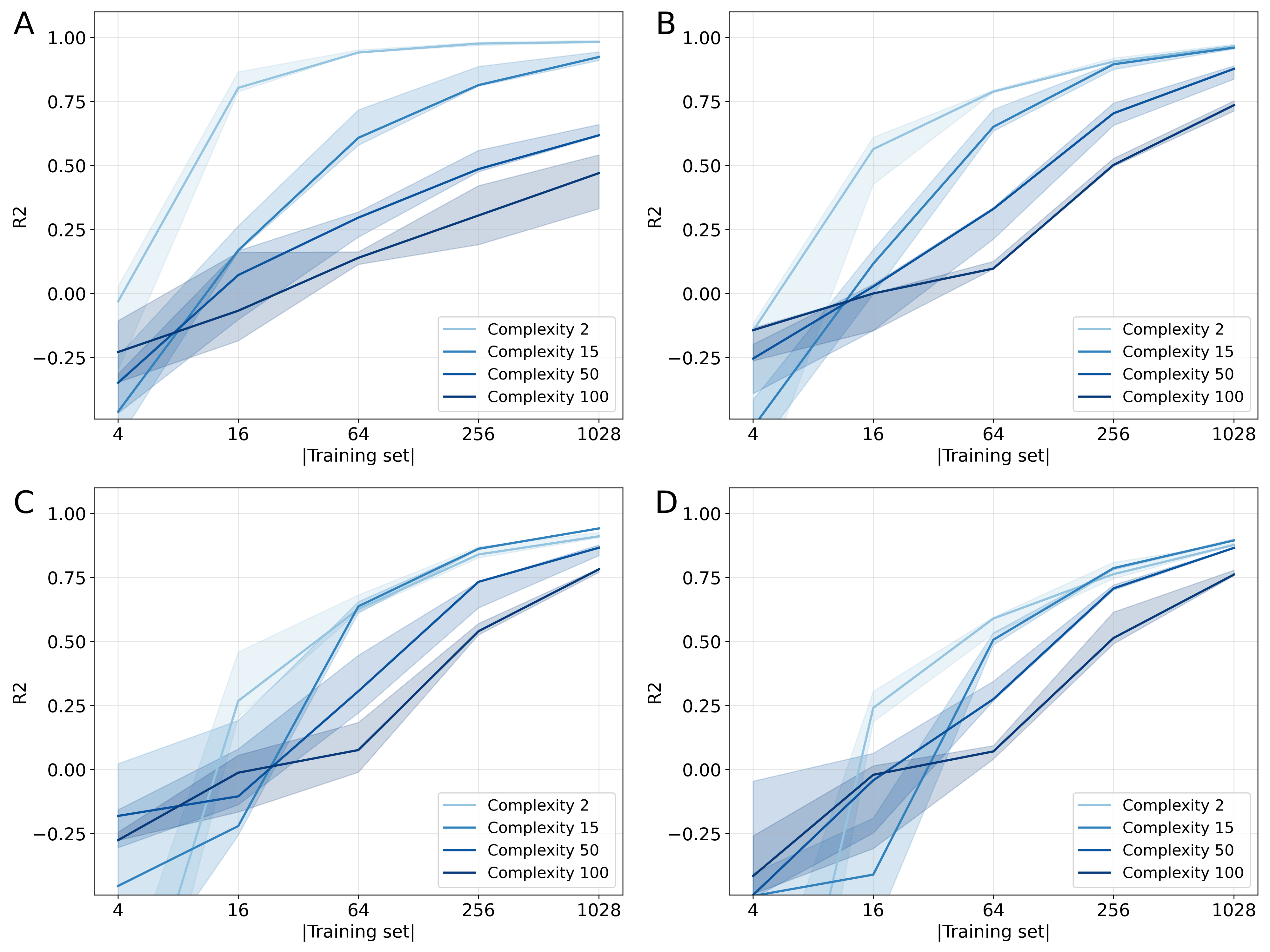}
    \caption{Correlation of $\mathcal{D}_{\text{seed}}$ size with function complexity. The plots show the coefficient of determination ($R^{2}$) on a test set for a support vector regressor as a function of training set size. Panels (A), (B), (C), and (D) correspond to input dimensions $d=2, 4, 6,$ and $8$, respectively. Different lines represent synthetic functions generated with varying $\mathcal{D}_{\text{seed}}$ size.}
    \label{fig:complexity}
\end{figure}

\paragraph{Function complexity.}
This complexity, in turn, is empirically shown to correlate with the amount of data required by a support vector regressor (implemented using scikit-learn with default settings) to achieve a target modelling accuracy. Figure \ref{fig:complexity} demonstrates this correlation: for different input dimensionalities ($d = 2, 4, 6,$ and $8$, corresponding to panels A, B, C, and D), we plot the coefficient of determination $R^{2}$ achieved on a fixed test set while varying the size of the training set, for functions generated with seed datasets of size 2, 15, 50, and 100. As the seed dataset size increases, the functions become more complex, requiring the model to use more training data to reach high accuracy.

\subsection{Multi-stage process}

To simulate an $N$-step cascade process (as defined in Section~\ref{sec:cascade} and Fig.~\ref{fig:block_diag}), we generate a sequence of $N$ functions, $f_1, \dots, f_{N}$, where the input of each step depends on the output of the preceding step.

For the initial step ($i=1$), the function $f_1$ accepts external input parameters $\boldsymbol{x}_1 \in [0, 1]^{d_1}$, defining the first hidden state as:
$$
\boldsymbol{h}_1 = f_1(\boldsymbol{x}_1).
$$
For all subsequent steps ($i>1$), the function $f_i$ accepts a concatenation of the current step's control parameters $\boldsymbol{x}_i \in [0, 1]^{d_i}$ and the hidden state output from the previous step, $\boldsymbol{h}_{i-1}$:
$$
\boldsymbol{h}_i = f_i([\boldsymbol{x}_i, \boldsymbol{h}_{i-1}]).
$$
The complexity of each component function $f_i$ is implicitly controlled by the size of its seed dataset $\mathcal{D}_{\text{seed}}$, as described in the previous section. A random generator initialised with a fixed seed is used for all data sampling and network initialisations, ensuring the complete process is fully reproducible.

To benchmark algorithms under realistic laboratory conditions, the framework supports the injection of stochasticity and information masking at various stages of the cascade. We distinguish between intrinsic process noise and extrinsic measurement uncertainty.

\paragraph{Process noise.}
Additive Gaussian noise can be injected directly into the latent process outputs $\boldsymbol{h}$ before they propagate to the next stage. This simulates inherent variability in the underlying physical or chemical transformation. The noisy process output is given by:
\begin{equation}
\boldsymbol{h}_i = f_i([\boldsymbol{x}_i, \boldsymbol{h}_{i-1}]) + \boldsymbol{\epsilon}^p,
\end{equation}
where $\boldsymbol{\epsilon}^{p} \sim \mathcal{N}(0, \sigma^2_{p}\boldsymbol{I})$
This noise propagates through the system, as $\boldsymbol{h}_{i}$ becomes the input for the subsequent step $f_{i+1}$, altering the trajectory of that specific sample in the cascade.

\paragraph{Measurements.}
To simulate partial observability, we apply a masking matrix $\boldsymbol{M}_i$ that selects a subset of the state $\boldsymbol{h}_i$ for observation, forcing the optimiser to infer the latent state of the system from partial data.
Furthermore, independent Gaussian measurement noise $\boldsymbol{\epsilon}^{m} \sim \mathcal{N}(0, \sigma^2_{m}\boldsymbol{I})$ can be added to these observations:
\begin{equation}
\boldsymbol{m}_i = \boldsymbol{M}_i \boldsymbol{h}_i + \boldsymbol{\epsilon}^{m}_i
\end{equation}
Unlike process noise, measurement noise affects only the data observed by the optimiser; it does not propagate to subsequent physical steps in the cascade.

\section{Additional results}
\label{app:additional}

\begin{figure}[H]
    \centering
    \includegraphics[width=0.99\columnwidth]{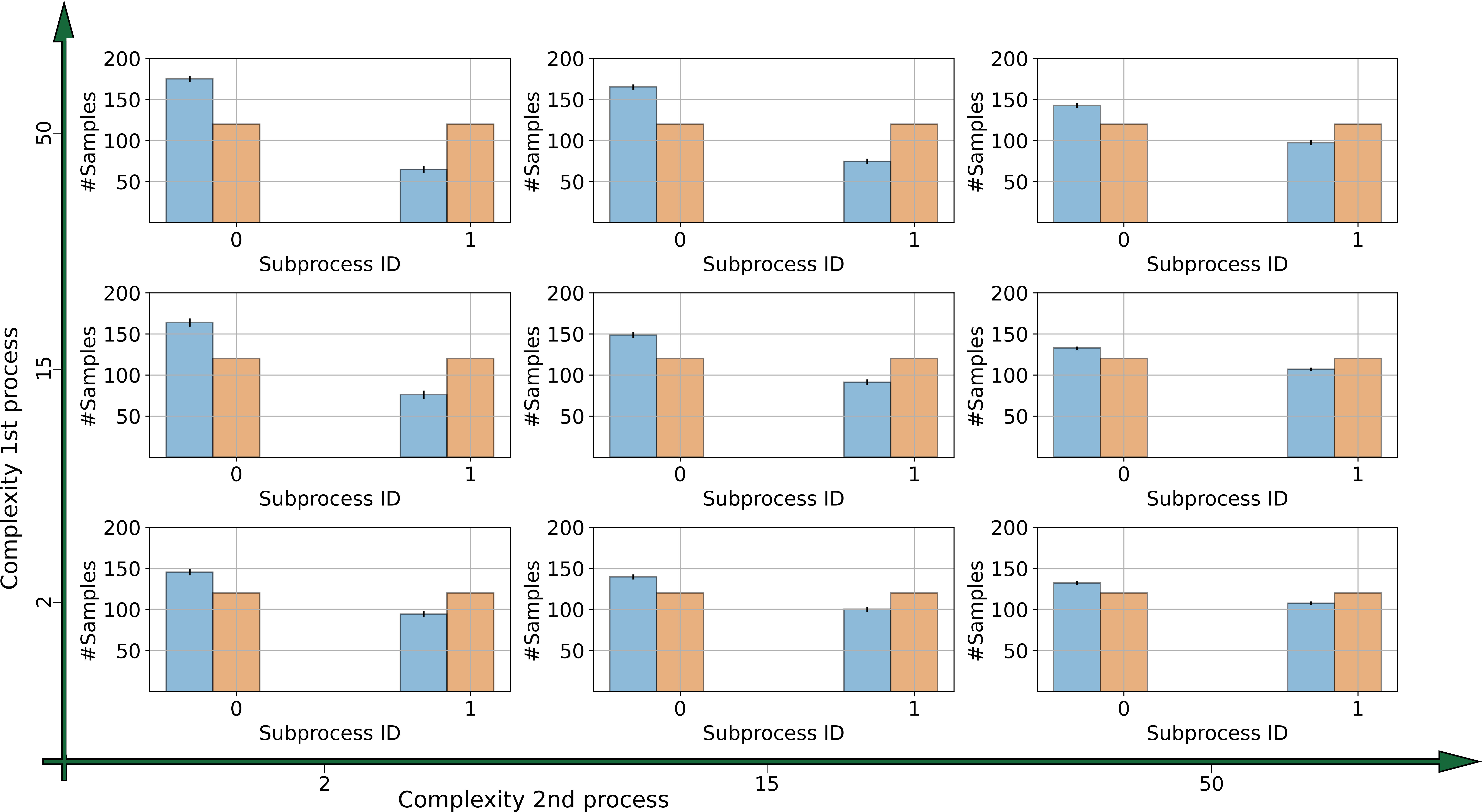}
    \caption{Sample counts across sub-processes. The bar charts display the number of samples collected for the first stage and the second stage, for MSBO (blue) and the baselines (orange). The grid arrangement corresponds to the complexity settings defined in Fig.~\ref{fig:complexity}, with rows representing increasing stage-one complexity and columns representing increasing stage-two complexity. The sampling distribution shifts dynamically according to the relative difficulty of each stage.}
    \label{fig:count_hist}
\end{figure}

\begin{figure}[H]
    \centering
    \includegraphics[width=0.75\columnwidth]{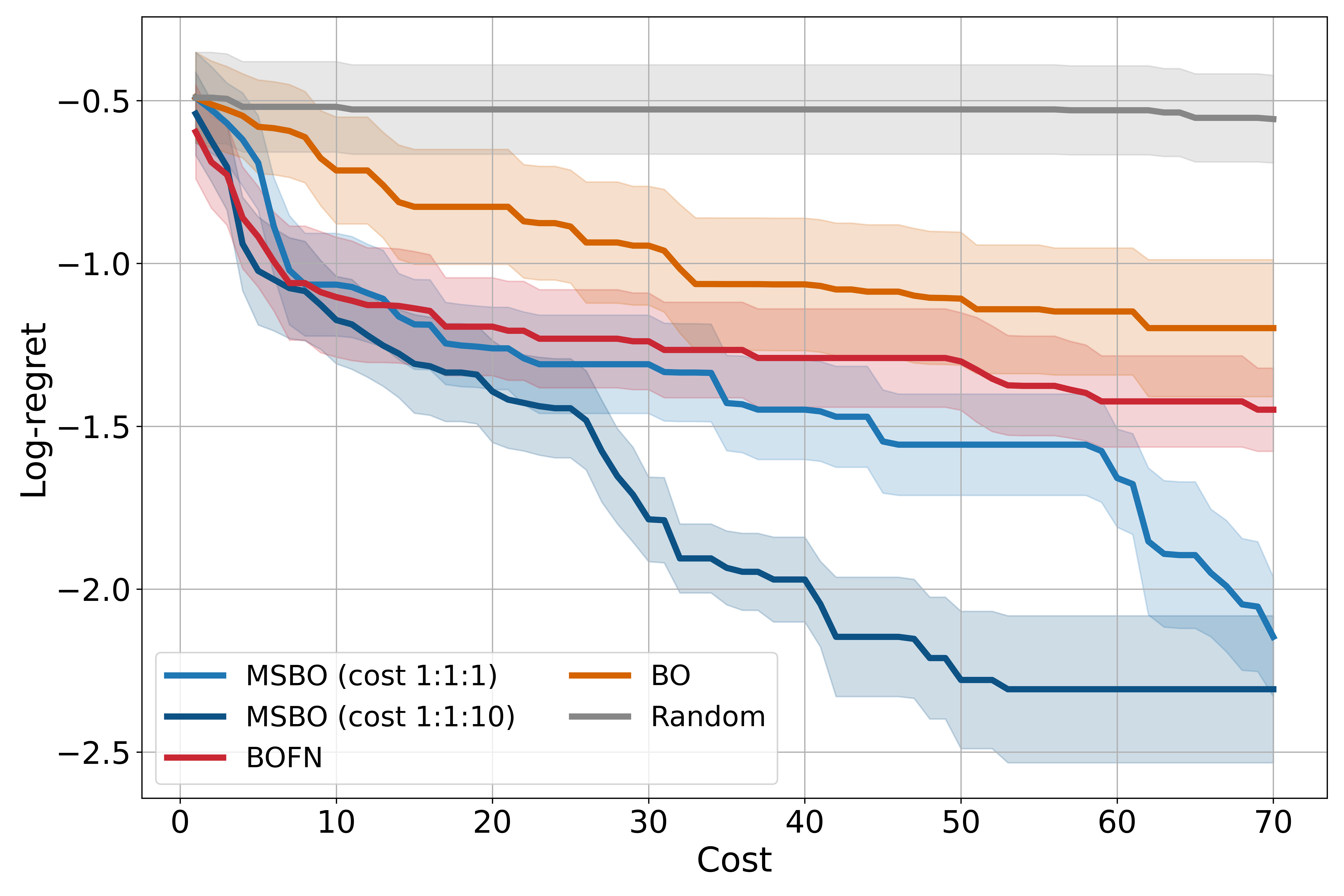}
    \caption{Performance on a three-stage cascade process. Aggregated log-regret plotted against cost for a synthetic three-stage process $f_{1}(\boldsymbol{x}_1) \rightarrow \boldsymbol{h}_1, f_{2}(\boldsymbol{h}_1, \boldsymbol{x}_2) \rightarrow \boldsymbol{h}_1, f_{3}(\boldsymbol{h}_2, \boldsymbol{x}_3) \rightarrow y$, with $\boldsymbol{x}_1 \in \mathbb{R}^4$, and $\boldsymbol{x}_2, \boldsymbol{x}_3, \boldsymbol{h}_1, \boldsymbol{h}_2 \in \mathbb{R}^2$, and $\mathcal{D}_seed$ size 15, 15, and 5, respectively. The curves display MSBO performance under uniform (1:1:1) and heterogeneous (1:1:10) cost ratios, compared to BO and BOFN baselines. The trajectories represent the aggregated results over 10 optimisation runs. Shaded regions indicate a 1 standard deviation interval. Cost and regret are considered unitless here.}
    \label{fig:three_steps}
\end{figure}

\begin{figure}[H]
    \centering
    \includegraphics[width=0.75\columnwidth]{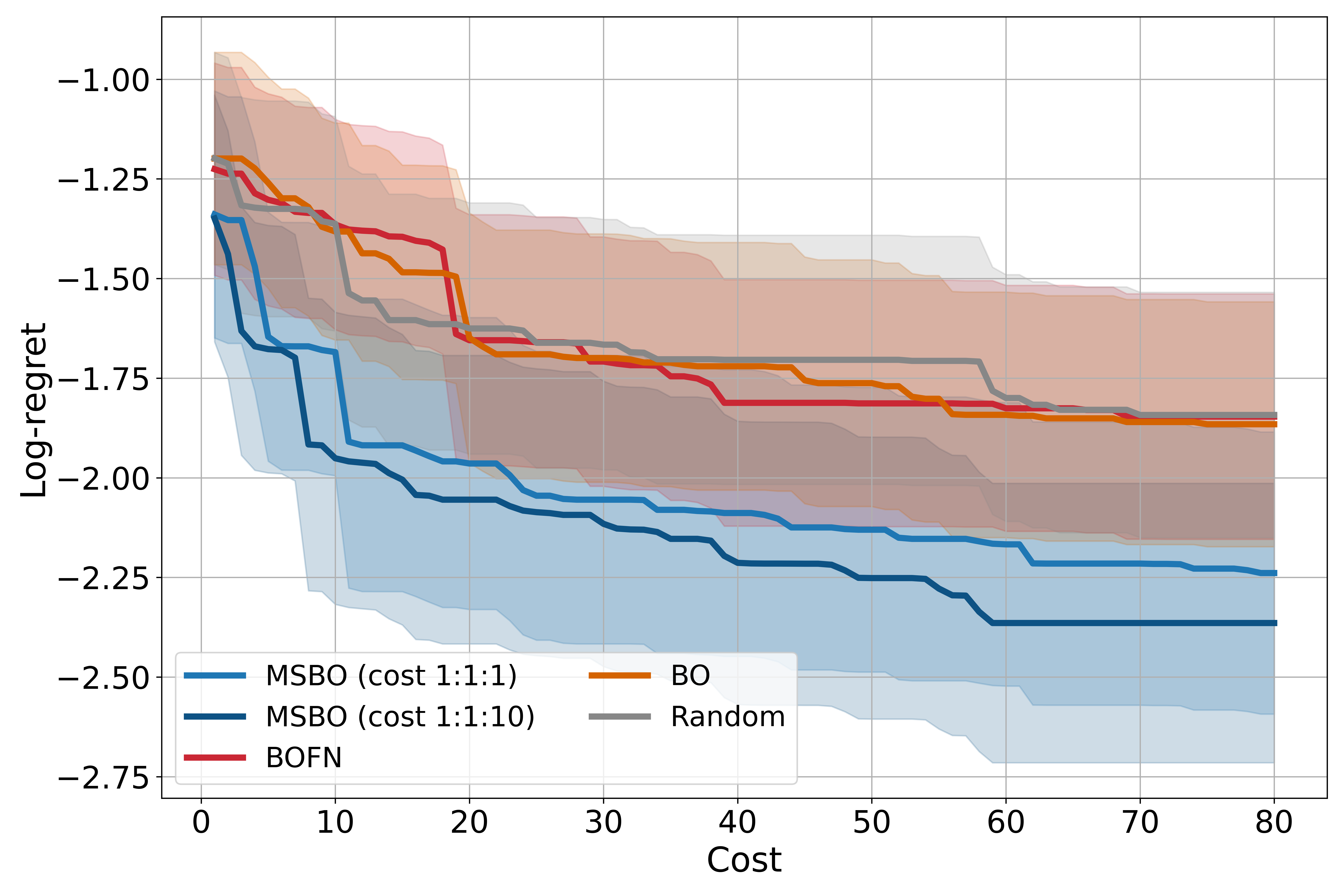}
    \caption{Performance on a partially observable three-stage cascade process. Aggregated log-regret plotted against cost for a synthetic three-stage process with controllable parameters $\boldsymbol{x}_1 \in \mathbb{R}^4$, $\boldsymbol{x}_2 \in \mathbb{R}^2$, and $\boldsymbol{x}_3 \in \mathbb{R}^1$. The latent process outputs are $\boldsymbol{h}_1 \in \mathbb{R}^4$, $\boldsymbol{h}_2 \in \mathbb{R}^2$, and $y \in \mathbb{R}^1$. Partial observability is simulated using a masking matrix $\boldsymbol{M}$ applied to the latent outputs, resulting in accessible intermediate observation dimensions of 2, and 1, respectively. The stage complexities ($\mathcal{D}_{seed}$ sizes) are 50, 15, and 2. The curves display MSBO performance under uniform (1:1:1) and heterogeneous (1:1:10) cost ratios, compared to BO, BOFN, and random baselines. The trajectories represent the aggregated results over 10 optimisation runs. Shaded regions indicate a 1 standard deviation interval. Cost and regret are considered unitless here.}
    \label{fig:three_steps_filtered}
\end{figure}

\begin{figure}[H]
    \centering
    \includegraphics[width=0.95\columnwidth]{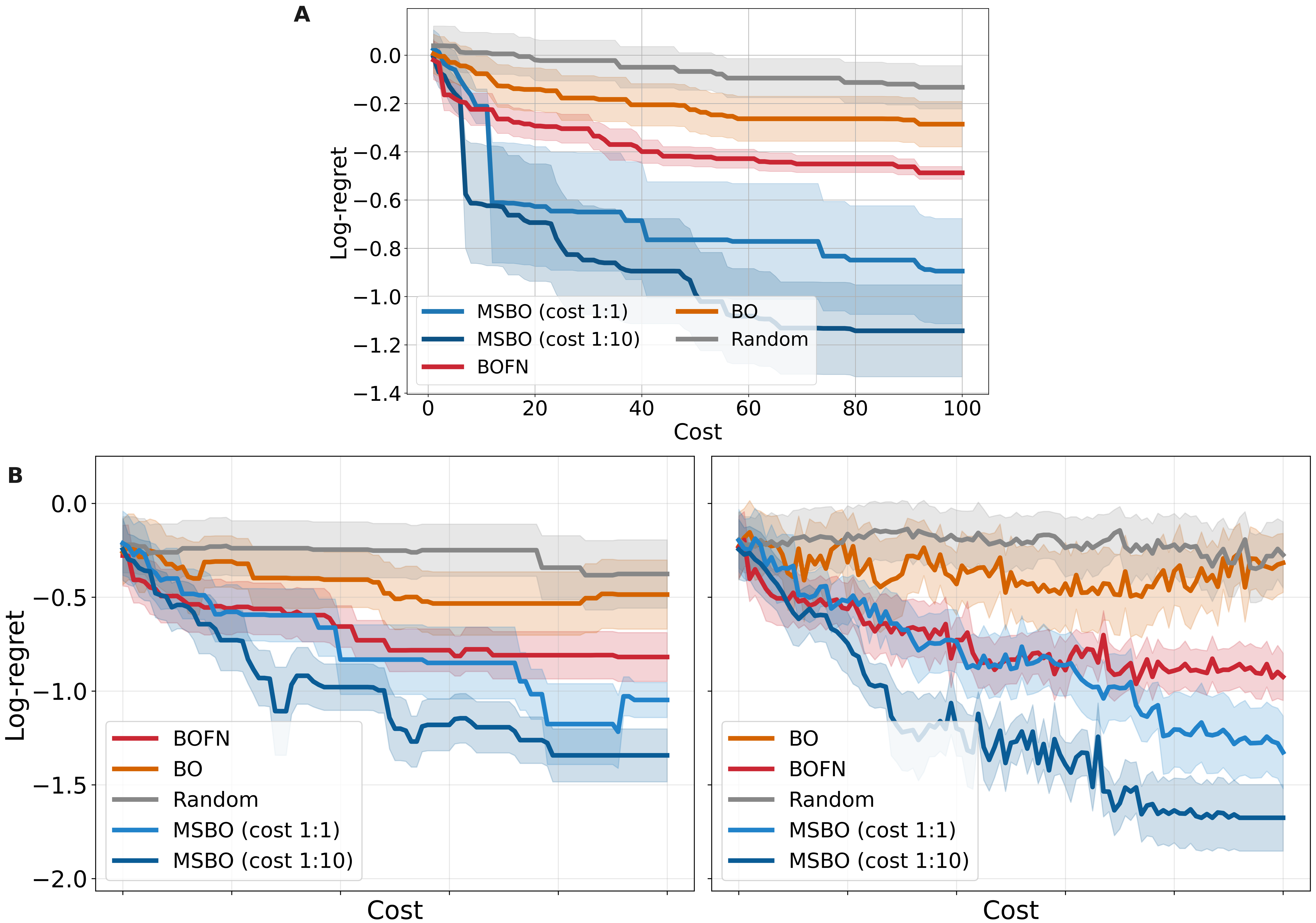}
    \caption{Assessment of optimisation robustness under process noise and the impact of model-based candidate selection. We evaluate a two-stage process with controllable parameters $\boldsymbol{x}_1 \in \mathbb{R}^4$ and $\boldsymbol{x}_{2} \in \mathbb{R}^2$, intermediate state $\boldsymbol{h}_1 \in \mathbb{R}^2$, complexities $(50, 2)$, and Gaussian process noise with standard deviations $(0.05, 0.1)$. (A) Simple regret calculated with respect to the noisy observations $(y_{opt} + 3 \sigma)$, where $\sigma$ is the estimated noise at the output. (B) Regret calculated with respect to $(y_{opt}$ of the denoised output. The left panel displays the denoised performance of the parameter set corresponding to the single best observed value (same parameters as in Panel A). The right panel displays the denoised performance of the parameter set selected at each iteration step by maximising the surrogate model's predicted mean among the collected samples.}
    \label{fig:noisy}
\end{figure}

\begin{figure}[H]
    \centering
    \includegraphics[width=0.9\columnwidth]{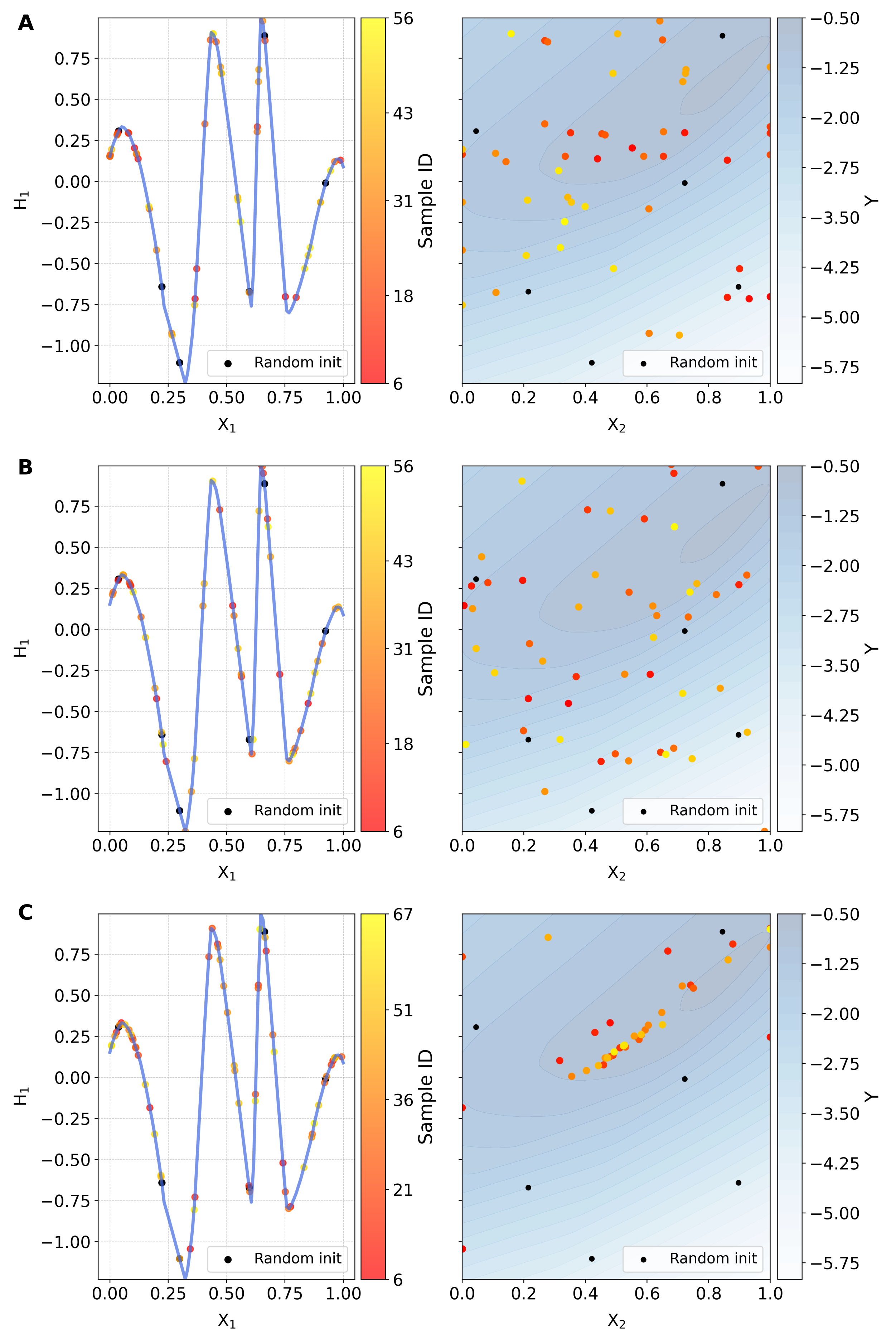}
    \caption{Representative sampling distributions for a two-stage process $f_{1}(x_1) \rightarrow h_1, f_{2}(h_1, x_2) \rightarrow y$. Panels (A), (B), and (C) correspond to standard BO, BOFN, and MSBO, respectively. The left subplots display the first-stage intermediate outputs $h_{1}$ (dots) plotted against the input $x_{1}$, overlaid on the true first-stage function. The right subplots show the locations of the second-stage evaluations within the $h_{1}\times x_{2}$ domain. Markers are colour-coded by the sample ID, with black dots representing random initialisation.}
    \label{fig:qualitative1}
\end{figure}

\begin{figure}[H]
    \centering
    \includegraphics[width=0.9\columnwidth]{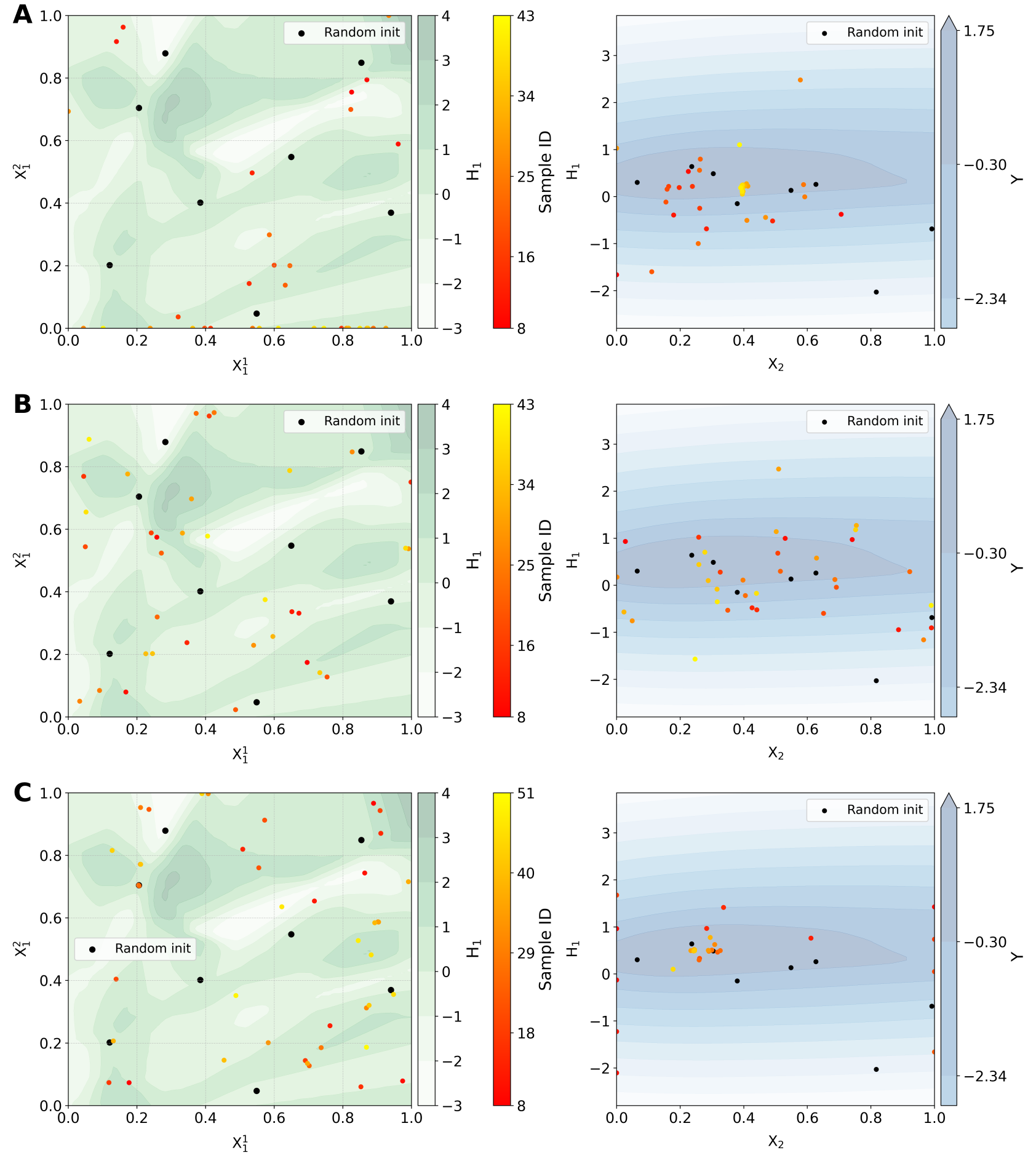}
    \caption{Representative sampling distributions for a two-stage process $f_{1}(x^1_1, x^2_2) \rightarrow h_1, f_{2}(h_1, x_2) \rightarrow y$. Panels (A), (B), and (C) correspond to standard BO, BOFN, and MSBO, respectively. The subplots display the sample locations in the 2D input space for the first stage (left) and the second stage (right), overlaid on the true function landscapes. Markers are colour-coded by the sample ID, with black dots representing random initialisation.}
    \label{fig:qualitative4}
\end{figure}